\documentclass{template/bmvc2k}

\usepackage{pgfplots}
\usepackage{tikz}
\usepackage{mathtools}

\usepackage[utf8]{inputenc} % allow utf-8 input
\usepackage[T1]{fontenc}    % use 8-bit T1 fonts
\usepackage{hyperref}       % hyperlinks
\usepackage{url}       
\usepackage[colorinlistoftodos,prependcaption,textsize=tiny]{todonotes}     % simple URL typesetting
\usepackage{booktabs}       % professional-quality tables
\usepackage{amsfonts}       % blackboard math symbols
\usepackage{nicefrac}       % compact symbols for 1/2, etc.
\usepackage{microtype}      % microtypography
\usepackage{xcolor}         % colors
\usepackage{amsmath}
\usepackage{amssymb}
\usepackage{pifont}
\usepackage{graphicx}
\usepackage{comment}
\usepackage{bbold}
\usepackage{subcaption}
\usepackage{amsthm}

\usepackage[colorinlistoftodos,prependcaption,textsize=tiny]{todonotes}

\newcommand{\bx}{\mathbf{x}}
\newcommand{\by}{\mathbf{y}}
\newcommand{\bobcaptionspace}{\vspace{0.3cm}}

\def\etal{\emph{et al}\bmvaOneDot}

\newcommand{\xmark}{\ding{55}}
\hypersetup{
  colorlinks   = true, %Colours links instead of ugly boxes
  urlcolor     = blue, %Colour for external hyperlinks
  linkcolor    = blue, %Colour of internal links
  citecolor   = red %Colour of citations
}

\pgfplotsset{colormap={mycolormap}{
        rgb255=(0,0,0)
        rgb255=(255,255,255)
    },
}

\title{Probabilistic Regression with Huber Distributions}

\addauthor{David Mohlin}{davmo@kth.se}{1}
\addauthor{Gerald Bianchi}{gerald.bianchi@gmail.com}{2}
\addauthor{Josephine Sullivan}{sullivan@kth.se}{3}

\addinstitution{
 Tobii AB / RPL, KTH
}

\addinstitution{
 Supervision during employment at Tobii
}
\addinstitution{
 RPL, KTH
}

\runninghead{Student, Supervisor, Prof}{BMVC Author Guidelines}

\begin{document}

\maketitle

\begin{abstract}
In this paper we describe a probabilistic method for estimating the position of an object along with its covariance matrix using neural networks.
Our method is designed to be robust to outliers, have bounded gradients with respect to the network outputs, among other desirable properties. To achieve this we introduce a novel probability distribution inspired by the Huber loss.
We also introduce a new way to parameterize positive definite matrices to ensure invariance to the choice of orientation for the coordinate system we regress over.
We evaluate our method on popular body pose and facial landmark datasets and get performance on par or exceeding the performance of non-heatmap methods.
Our code is available at \href{https://github.com/Davmo049/Public_prob_regression_with_huber_distributions}{github.com/Davmo049/Public\_prob\_regression\_with\_huber\_distributions}
\end{abstract}

\section{Introduction}

Estimating positions of objects is a well studied topic, due to its many applications. It is for example used for facial landmark estimation \cite{kumar2020luvli}, autonomous driving \cite{Girshick_2015_ICCV,Chen_2016_CVPR,Mousavian_2017_CVPR} and body pose estimation \cite{Felzenszwalb03pictorialstructures,Toshev_2014_CVPR,Sarafianos_2016_CVIU,sun2017compositional}.
Regression can also appear as a component of more complicated systems such as predicting offsets of a bounding box relative to an anchor point for object detection \cite{yolov3}.
Estimating uncertainties associated with these estimated positions has applications for example in time filtering, such as Kalman filters. Uncertainties can also be used for task specific problems, for example an autonomous vehicle should be able to come to a stop before it enters the region where an object is likely to be, not before it gets to the most likely position of the object.

% robust section
Defining a loss function is a crucial step in designing a neural network. Robustness to outliers is an important property of the loss in order to reduce sensitivity to large errors. Robust loss functions for neural networks have been shown to improve performance of the model. In \cite{Feng_2018_CVPR} they show that minimizing L1 and smooth L1 loss yields higher performances than the standard L2 loss. Motivated this they introduce the Wing loss, which decreases the impact large errors have. An extension of the Wing loss was presented in \cite{Wang_2019_ICCV} in the context of heatmap regression. In \cite{barron19cvpr}, the author presents a generalization of several common loss functions with a robustness parameter which is automatically tuned during the training step. This approach allows the optimization to determine how robust the loss should be depending on the training data

Many of the current approaches to regress positions can be divided into two categories: heatmap based methods and direct regression. Heatmap based methods estimate a heatmap of where the object could be over a quantized set, roughly corresponding to pixels in the image. This heatmap is then converted into a position through various heuristics such using the expected position, the most likely position or something similar. Direct regression does not introduce this intermediate representation and instead predict a vector in $\mathbb{R}^N$ directly from the latent representation of the network.
Direct regression has been applied on for example depth estimation \cite{barron19cvpr} and estimation of facial landmarks \cite{Feng_2018_CVPR}.

In general heatmap methods are more complicated, requiring heuristics for how to construct the loss and how to extract a position out of the heatmap. State of the art methods often also introduce other complex components such as using a cascade of deep neural networks \cite{kumar2020luvli,newell16eccv_stackedhourglass,tang20_dunet} or multi-resolution networks \cite{Sun_2019_CVPR}.

While heatmap methods are currently the prevalent approach, we think that direct regression methods are still attractive for solving the problem of position estimation for three reasons.
Firstly, direct regression methods do not quantize the output space into bins, which results in quantization errors and scales poorly to higher dimensions.
Secondly, direct regression directly output the coordinates of the object, removing the need for a heuristic which convert heatmaps to coordinates. 
Finally, direct regression do not require complex network architectures to produce heatmaps, instead any standard network backbone such as resnet, inception, mobilenet or squeezenet can be used.

We evaluate this method on popular facial landmark (WFLW) and body pose (MPII) datasets. The results show that our method outperforms existing regression methods but gives slightly lower performance than the state of the art for heatmap based methods.

In this paper we introduce the Huber $L_2$ distribution a novel probability distribution, parameterized by a mean position and a covariance matrix. We fit neural networks to predict this prediction by  minimizing the negative log likelihood between the predictions and annotations. We design the method to ensure the following desirable properties: i) unimodality of the distribution ii) using a distribution with exponential tail behaviour to make the method robust to outliers, iii) make the method invariant to the orientation of the coordinate system we regress over, iv) have bounded gradients to avoid too large parameter updates in gradient descent v) make the method loss have bounded hessians and make the loss convex for a  region which we argue covers all reasonable outputs.

\begin{figure}[t!]
  \begin{center}
\includegraphics[width=.8\linewidth]{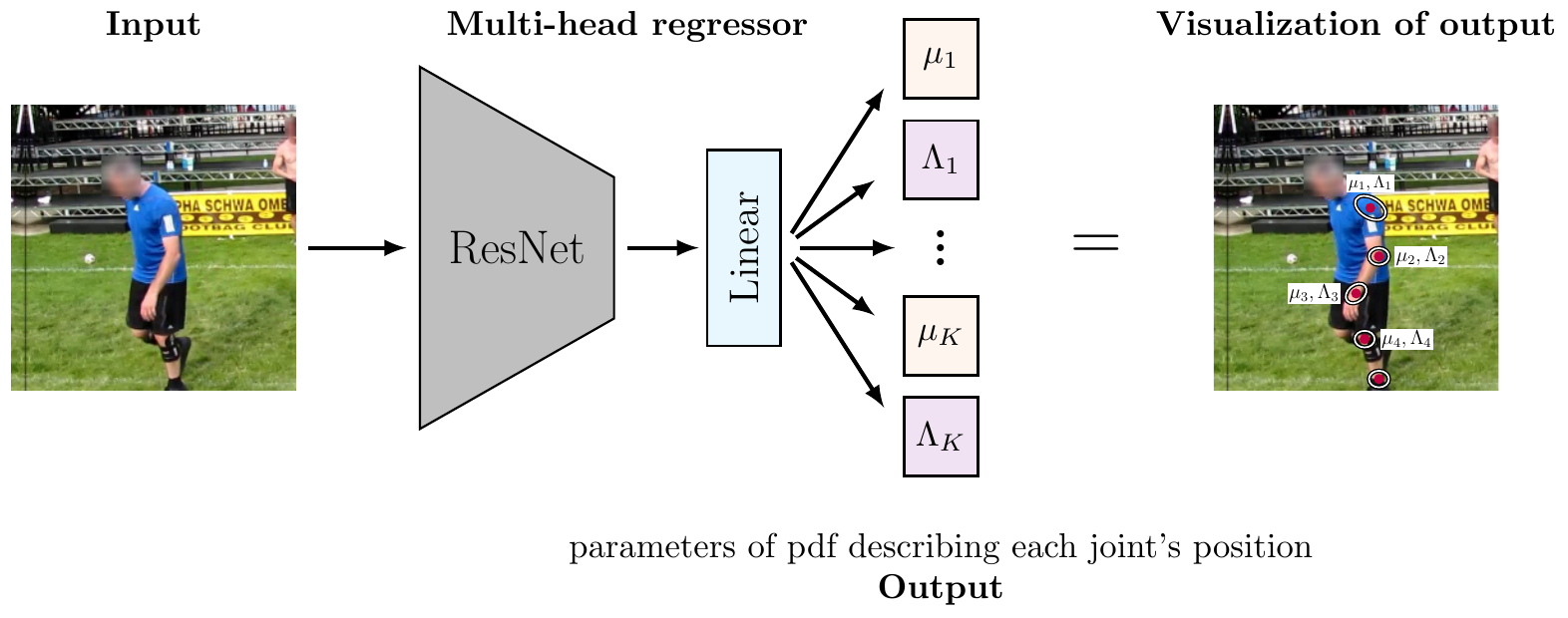}
\end{center}
\caption{\textbf{Overview of our simple regression architecture.} In this paper we introduce a simple regression method that produces a probabilistic estimate of a $d$-dimensional vector. An input image is fed into any standard backbone network architecture with an additional final linear layer that outputs the parameters, mean $\mu$ and \textit{precision} $\Lambda$, of the \textit{$L_2$ multivariate Huber distribution} describing the likely values of the vector. This Huber distribution is introduced to allow robust estimation from noisy landmark datasets common in computer vision. Above depicts our method adapted to predict the location of body joints.}

\label{fig:splash_fig}
\end{figure}
\section{Related Work}

Multivariate regression and estimating 2D and 3D position has been extensively studied in computer vision. Both with classic machine learning and deep learning techniques.

The state of the art methods for pose estimation on the body pose dataset MPII \cite{andriluka14cvpr} have been heatmap based since 2014 \cite{tompsonjoint, tompson2015efficient, newell16eccv_stackedhourglass, yang2017learning, Tang_2018_ECCV, su2019cascade, bin2020adversarial}.
Some innovations introduced since then are multi-stage losses \cite{carreira2016human}, the hourglass network architecture \cite{newell16eccv_stackedhourglass} and exploiting additional training datasets \cite{wu2017leveraging, su2019cascade}.
For the facial landmark dataset WFLW \cite{wayne2018lab} heatmap based methods are also popular for achieving high performance \cite{qian2019aggregation, Wang_2019_ICCV, kumar2020luvli}.

An intermediate heatmap representation is not the basis of all position estimation methods. Carreira \etal \cite{carreira2016human} use regression to predict landmark positions, these predictions are then converted into heatmaps and concatenated to the original image to be sent through a second network for the final prediction.
Feng \etal \cite{Feng_2018_CVPR} introduce the Wing loss to focus more on small and medium size errors during the training when predicting facial landmarks, effectively making their method less affected by outliers.
Sun \etal \cite{sun2017compositional} modeled the position of body joints as a tree with the pelvis as a root and the position of a child joint being defined by an offset from its parent.

%covar
For the problem of predicting uncertainty for position it is common to construct a covariance matrix from the network output.
The covariance matrix is symmetric positive definite by definition. One method to do this to estimate a diagonal covariance matrix \cite{barron19cvpr}, by applying a mapping from $\mathbb{R}$ to $\mathbb{R}^{+}$ on each value, such as a softplus function, it is possible to guarentee positive definiteness. A problem with this approach is that diagonal matrices is only a subset of all positive definite matrices.. Another common method is to predict parameters for a decomposition of the matrix using neural networks, followed by a reconstruction of the covariance matrix. Common decompositions to predict parameters of are the LDL decomposition\cite{Liu_2018_ICRA} or the Cholesky decomposition \cite{Gundavarapu_2019_CVPR_Workshops, kumar2020luvli}.
\section{Method}
For an input, $\bx \in \mathbb{R}^D$, we want to predict $\by \in \mathbb{R}^d$ in a probabilistic fashion. We achieve this by training a neural network that outputs the parameters of a probability distribution for $\by$ over $\mathbb{R}^d$. In this section we introduce a novel distribution for this purpose and from it derive the loss we use for training our network. This loss and its parameterization from network output is designed to be invariant to the choice of orientation of the coordinate system we do regression in, have bounded gradients, is convex for the set of precision matrices with eigenvalues larger than a threshold corresponding to the smallest precision value deemed reasonable for the specific regression task and has bounded Hessians for the same set. These properties are proven in supplementary material \ref{sec:supp_loss}. Since one of the parameters of our distribution is a symmetric positive definite matrix, we also describe the procedure used to map the network's output to a symmetric positive definite matrix.

\subsection{The $L_2$ multivariate Huber distribution}
\label{sec:method_multivar_huber}
We define the $L_2$ \textit{multivariate Huber distribution} for $\mathbf{y} \in \mathbb{R}^d$ as
\begin{equation}
  p(\mathbf{y} \mid \mu, \Lambda, \delta) \propto \exp\left(-h_{\delta}\left(\|\Lambda^{1/2}(\mathbf{y}-\mu)\|_2\right)\right)
  \label{eqn:l2_huber_prob}
\end{equation}
where $\Lambda \in \mathbb{R}^{d\times d}$ is a symmetric positive definite matrix, $\mu \in \mathbb{R}^d$ and
\begin{equation}
 h_{\delta}(y) = \begin{cases} 
y^2/2 & \text{if $|y| \le \delta$} \\[3pt]
\delta(|y|-\delta/2) & \text{otherwise}
\end{cases}
\end{equation}
The function $h_{\delta}$ is the well known Huber function parametrized by $\delta \in [0, \infty)$. Intuitively $\mu$ is the mean position and $\Lambda$ is the precision multiplied by a constant, see supplementary material \ref{sec:supp_variance_of_huber} for details.  Our multivariate Huber distribution is similar, but not identical to the multivariate Huber distribution defined in \cite{aravkin2010robust}. In the latter the Huber loss is applied independently to each dimension of the vector as opposed to in our distribution (equation (\ref{eqn:l2_huber_prob})) where the Huber loss is applied on the $L_2$ norm. Our distribution has a tail density which is $\mathcal{O}(\exp(-\|y\|_2))$, compared to a Gauss distribution which has $\mathcal{O}(\exp(-\|y\|_2^2))$. This makes maximum likelihood estimators of $\mu$ and $\Lambda$ less dependent on outliers, compared to a Gauss distribution. This is an important property for robust estimation in the presence of mislabelings and heavy-tailed noise. See the section \ref{sec:huber_vis} for visualizations of the distribution.

When we estimate the parameters of the Huber distribution we use the negative log-likelihood of the distribution as the training loss. A slightly different parametrization of equation (\ref{eqn:l2_huber_prob}) where we instead estimate $A = \Lambda^{1/2}$, $\nu = \Lambda^{1/2}\mu$ gives this loss nice mathematical properties. The multivariate Huber distribution becomes, including the normalising constant and dropping the subscript in $\|\cdot\|_2$:
\begin{equation}
  p_{\text{\tiny huber}}(\mathbf{y} \mid \nu, A, \delta) = \frac{|A|}{c_d(\delta)} \exp\left(-h_{\delta}\left(\|A\mathbf{y}-\nu)\| \right)\right)
  \label{eqn:l2_huber_prob_param}
\end{equation}
where $|\cdot|$ denotes the determinant and $c_d$ is a constant depending only on the dimensionality of $\by$ and $\delta$. See supplementary  \ref{sec:supp_normalizing_factor_and_var} for details about $c_d$. Note that in our experiments we set $\delta$ to be constant.

\subsection{Loss based on the $L_2$ multivariate Huber distribution}

Assume we have labelled training data, $\mathcal{D}$, where an example $(\bx, \by) \in \mathcal{D}$ has input $\bx \in \mathbb{R}^D$ and a corresponding position vector $\by \in \mathbb{R}^d$. Our goal is to train a neural network that will map an input $\bx$ to a probabilistic estimate of its position. We learn a function, $f$, defined by parameters $\Theta \in \mathbb{R}^p$ and encoded as a neural network that outputs a vector of length $(d+d(d+1)/2)$ given input $\bx$. We then apply a function $q$ to the output vector to return the parameters $\nu$ and $A$ of a $L_2$ multivariate Huber distribution that is  
\begin{align}
q(f(\bx, \Theta)) = (\nu_{\bx, \Theta}, A_{\bx, \Theta}) \quad \text{where}\quad q: \mathbb{R}^{d+d(d+1)/2} \rightarrow \mathbb{R}^d \times \mathbb{R}^{d\times d}   
\end{align}
%where $\Theta \in \mathbb{R}^p$ denotes the network's parameters and
\begin{comment}
\begin{align}
  f: \mathbb{R}^D \times \mathbb{R}^p  \rightarrow \mathbb{R}^{d+d(d+1)/2}
  \quad \text{and}\quad q: \mathbb{R}^{d+d(d+1)/2} \rightarrow \mathbb{R}^d \times \mathbb{R}^{d\times d}   
  %% f: \mathbb{R}^D \times \mathbb{R}^p \rightarrow \mathbb{R}^{d} \times \mathbb{R}^{d \times d}
  %% \quad
  %\text{with } f(\bx, \Theta) = (\nu_{\bx, \Theta}, A_{\bx, \Theta})   
\end{align}
As a shorthand we define $q(f(\bx, \Theta)) = (\nu_{\bx, \Theta}, A_{\bx, \Theta})$
where $\Theta \in \mathbb{R}^p$ denotes the network's parameters. 
\end{comment}
The parameters, $\Theta$, can be found by maximimizing the likelihood of the training data w.r.t. the multivariate Huber distribution of equation (\ref{eqn:l2_huber_prob_param}) or equivalently minimizing the negative log-likelihood of the training data:
\begin{align}
  \underset{\Theta}{\arg \max}\; \prod_{(\bx, \by) \in \mathcal{D}} p_{\text{\tiny huber}}(\by \mid \nu_{\bx, \Theta}, A_{\bx, \Theta})
  &=
  \underset{\Theta}{\arg \min}\; \sum_{(\bx, \by) \in \mathcal{D}} -\log\left(p_{\text{\tiny huber}}(\by \mid \nu_{\bx, \Theta}, A_{\bx, \Theta})\right)
  \label{eq:max_prob_min_loss}
\end{align}
When dropping the $\Theta$ subscript to reduce the notation clutter in equation (\ref{eq:max_prob_min_loss}) and using the fact that $\delta$ is constant we get the loss
\begin{align}
  \mathfrak{L}(\mathbf{y}, \nu_{\bx}, A_{\bx}) = -\log(|A_{\bx}|) + h_{\delta}(\|A_{\bx} \mathbf{y} -\nu_{\bx})\|)
  \label{eq:loss}
\end{align}

In our experiments we regress for multiple keypoints simultaneously. We do this by increasing the number of outputs of the network proportional to the number of keypoints and sum the losses, equation (\ref{eq:loss}), for each keypoint into one total loss. In a probabilistic framework this corresponds to the assumption that the position of each keypoint is independent.

\subsection{Predicting a symmetric positive definite matrix}

\begin{comment}
When we design the function $q$ we need it to both span all output parameters which we want to be able to modes as well as conform to the constraints imposed on the distributions parameters.
To model $\nu$ we need to output a $\mathbb{R}^d$ vector. This can be achieved by for example
\end{comment}
The function $q$, mapping the network's output to the distribution's parameters, should both span all possible output parameters as well as fulfill the parameter's constraints.
For the mean vector $\nu \in \mathbb{R}^d$, we can simply let $\nu$ correspond to the first $d$ numbers output by $f$.
%\begin{equation}
%\nu_i = f(\bx, \theta)_i \quad \forall i \in \{1,2, \ldots, d\}
%\end{equation}
Generating the matrix $A_{\bx}$ is trickier as it needs to be symmetric positive definite. We also want to ensure the procedure we use to construct $A_{\bx}$ does not bias this matrix to have certain eigenvectors for certain eigenvalues. This motivates the following approach. 

Let the vector $\mathbf{v}$ correspond to the last  $d(d+1)/2$ numbers output by $f$.
%Start by constructing a vector $\mathbf{v}$ 
%\begin{equation}
%v_i = f(\bx, \theta)_{d+i} \quad \forall i \in \{1,2,\cdots , d(d+1)/2\}
%\end{equation}
From $\mathbf{v}$ we can create a symmetric matrix $B$ with a bijection $\pi$
\begin{equation}
    B_{i,j} = \begin{cases}
    v_{\pi(i,j)} / \sqrt{2} &\text{ if } i > j \\
    v_{\pi(j,i)} / \sqrt{2} &\text{ if } i < j \\
    v_{\pi(i,i)} &\text{ if } i = j
    \end{cases}
\end{equation}
where
\begin{equation}
    \pi: \left\{(i,j) \mid i,j \in \left\{1, \ldots, d \right\} \;\&\; i \le j\right\} \longrightarrow \left\{1, \ldots, d(d+1)/2\right\}
\end{equation}
Next we perform an eigenvalue decomposition of $B = V^T \text{diag}(\lambda_1, \ldots, \lambda_d)\, V$. $B$ being symmetric and real ensures $V$ is orthonormal (ON) and each eigenvalue $\lambda_i$ is real. We then construct a positive definite matrix by applying a function $g:\mathbb{R} \rightarrow \mathbb{R}^{+}$ on each eigenvalue independently.

\begin{equation}
  g(\lambda) = \begin{cases}
    \lambda & \text{if $\lambda > \theta$}\\[2pt]
    \theta \exp\left(\lambda/\theta - 1 \right) & \text{otherwise}    
  \end{cases}
  \label{eqn:evalues_pos}
\end{equation}
The value of $\theta$ corresponds to the smallest reasonable precision for the task. For example when doing regression on an image one should not need to be able to output distributions with a standard deviation larger than the size of the image.

We can then construct our output matrix as
\begin{equation}
 A = V^T \text{diag}(g(\lambda_1), \ldots, g(\lambda_d))\, V
\end{equation}
This procedure is not biased to output certain eigenvectors for certain eigenvalues if $\mathbf{v}$ is not biased toward certain directions. This is because the mapping to create $B$ is an isometry between $\mathbb{R}^{d(d+1)/2}$ w.r.t the $L_2$ norm and the set of symmetric matrices with respect to the Frobenius norm. Since the Frobenius norm only depends on the eigenvalues, not the eigenvectors this means that the eigenvectors of $A$ and $B$ will be unbiased.

It is not completely straightforward to do backward propagation through the above sequence of steps, but the analytical expression and proof of its correctness is available in supplementary material \ref{sec:sup_mat_diag_remapping}.

\section{Experiments}
We evaluate the effectiveness of our probabilistic Huber loss and approach on the problem of keypoint prediction for images. The two datasets we use are the facial landmark dataset WFLW \cite{wayne2018lab} and the 2D body keypoint dataset MPII \cite{andriluka14cvpr} where the goal is to estimate the 2D position of each landmark in the image. The facial landmark dataset demands high precision while the human pose dataset has a large degree of variation.
We run our method 5 times with different random seeds and report the mean performance of the runs.

\subsection{Implementation details}
\paragraph{Preprocessing}
\label{sec:preprocessing}
Both MPII and WLFW provide bounding boxes for each person/face considered as well as the annotated keypoints corresponding to each face. The input to our network at test time is constructed by creating a square crop centered at the center of the bounding box. The length of the cropped region is proportional to the distance between the bounding box corners and its center. The crop is then scaled, keeping the aspect ratio intact, to the network's expected input size. 

For training time we do the same except we also use standard data augmentations such as mirroring the image, using random rotations, scaling, translations and perspective distortions. All of these operations can be combined into an affine transform. We construct the input image sampling pixels using bilinear interpolation and edge replication using this transform. By only sampling pixel values once we avoid creating excessive blur. We use the same affine transform to compute where each annotated landmark will be in the input image.

Finally we apply an affine normalization for each landmark such that each normalized landmark is zero mean with identity covariance across all training samples.

\paragraph{Test time data-augmentations}
\label{sec:huber_fusion}
Test time augmentations are frequently applied to improve performance on MPII and WFLW \cite{bin2020adversarial, su2019cascade, Tang_2018_ECCV, yang2017learning, newell16eccv_stackedhourglass}. We do this by combining the outputs of our model for a non-augmented input and a mirrored input. We convert the predictions for the mirrored input to the same coordinate system as the non-augmented output by using the affine transforms used for preprocessing.

The standard way to combine test time augmentations is by averaging the predictions. Since our method outputs probability distributions, we can fuse our predictions using the maximum likelihood (ML) point. Empirically the two methods performed similarly, this could be because a mirrored and non-mirrored input produces similar covariance matrices. Finding the ML point for multiple independent Huber can be done by using Majorize/Minimize (MM) of quadratic functions for quick and guaranteed convergence. See Supplementary material \ref{sec:supp_huber_fusion} for details.

\paragraph{Other implementation details}

For our experiments we use the ResNet family of convolutional networks\cite{he2016deep} as our backbone network. We use an input size of $224 \times 224$ pixels.

For this input we get an output with a spatial resolution of $7 \times 7$. Conventionally this spatial resolution is then reduced to  $1 \times 1$ by average pooling. Pooling is often motivated by a desire to have invariance to small translation, scale and/or rotation changes. However, this property is undesirable for regression and we replace this average pooling with channel wise convolutions whose parameters are learned. This change improved performance by 0.24 NME for WFLW, see supplementary material \ref{sec:supp_extra_tables} for corresponding tables.

For the regression head we use a linear mapping from the latent space to a $5K$ dimensional output, where $K$ is the number of keypoints we want to estimate and 5 is the number of parameters we need to parameterize our  $L_2$ multivariate Huber distributions for two dimensions.
The $\delta$ parameter of the loss is set to be 1.

\subsection{Evaluation metrics}
\label{sec:eval_metrics}
There are standard evaluation metrics associated with the datasets WFLW and MPII. For WFLW it is  the \textit{Normalized Mean Error} (NME) between the predicted and ground truth landmark coordinates where the normalization is performed w.r.t. the interoccular distance.  
 The standard evaluation metric for MPII is PCKh(@0.5). The metric measures the percentage of the predicted keypoint locations whose distance to its ground truth location are within 50\% of the head segment's length. We now review of how we quantitatively evaluate our probability distributions. 
 
\paragraph{Quantitative evaluation of uncertainty and error} \label{sec:eval_metric_uncert_and_error}
We investigate whether the estimated precision matrices for test samples are correlated with the actual prediction error. To achieve this we use a similar approach to \cite{kumar2020luvli}. First for each test sample we compute its expected error from the predicted covariance. Next we group samples with similar expected errors into bins. For the presented plots we use a bin size of 734.  Finally we plot the average expected error against the average empirical error for each bin.

\section{Results}

\subsection{Performance relative to other methods}

Table \ref{tab:sota_comparison} reports the performance of our best performing networks for WFLW and MPII compared to those of recent high-performing approaches which involve landmark heatmaps in some form or other and those which do not. Our best-performing networks have a ResNet101 backbone architecture, are trained using our probabilistic Huber loss with a $\nu$ parametrization for 200 epochs and use test-time augmentation with our probabilistic fusion. For WFLW our best performing network ranks $\sim$4th, w.r.t. the NME metric, of all published methods and is the best performing method which directly regresses from a compact encoding of the input image. For MPII our results are respectable, given our streamlined approach, but significantly below state of the art performance. Once again our approach is best amongst single stage regression from a compact representation. 

\begin{table}[htbp]
  \caption{Comparison of our regression approach to state of art methods on WLFW and MPII. Method marked with $\dagger$ construct heatmaps of predictions as additional input for multi stage regression. Entries marked with * use additional data.}
  \bobcaptionspace
  \begin{subtable}[t]{0.3\textwidth}
    \subcaption{\normalsize WFLW}
    \raggedright
    \begin{tabular}{@{}lcr}
      \toprule
      Method & Heatmaps & NME\\      
      \midrule
CFSS\cite{zhu2015face} & \xmark & 9.07 \\
         DVLN\cite{wu2017leveraging} & \xmark & 10.84\\
         LAB\cite{wayne2018lab} & \checkmark & 5.27  \\ % predict heatmap, fusion heatmap with feature map for normal regression.
         Wing\cite{Feng_2018_CVPR} & \xmark & 5.11  \\ % seems like normal regression
         DeCaFa\cite{dapogny2019decafa} & \checkmark & 4.62  \\  % predict heatmap, use expected position given softmax prob. dist to get differentiable coordinate estimate.
         AVS\cite{qian2019aggregation} & \checkmark & 4.39 \\ % Probably, this paper uses SAN which might be the worst code i've seen
         AWing\cite{Wang_2019_ICCV} & \checkmark & \textbf{4.36} \\
         LUVLi\cite{kumar2020luvli} & \checkmark & 4.37  \\
         \midrule
         Ours & \xmark & 4.62  \\
         \bottomrule
    \end{tabular}
  \end{subtable}%
  \hfill
  \begin{subtable}[t]{.6\textwidth}
    \subcaption{\normalsize MPII}
    \raggedleft
    %\centering
    \begin{tabular}{@{}l c c@{}}
      \toprule
      Method & Heatmaps & PCKh(@0.5) \\
      \midrule
      Tompson \etal \cite{tompsonjoint} & \checkmark  & 79.6 \\
      Tompson \etal \cite{tompson2015efficient} & \checkmark & 82.0 \\
      Newell \etal \cite{newell16eccv_stackedhourglass} & \checkmark & 90.9 \\ % hourglass heatmaps
      Yang \etal \cite{yang2017learning} & \checkmark & 92.0 \\ % Hourglass heatmaps
      %Tang ECCV18 \cite{Tang_2018_ECCV} & \checkmark & 92.3 \\ % Hourglass heatmaps for different "level" of concepts e.g. arm, elbow etc.
      %Su ArXiv19* \cite{su2019cascade} & \checkmark & 93.9 \\ % Heatmaps
      %Bulat FG2020* & 94.1 & \xmark\\ % Heatmaps
      Bin \etal* \cite{bin2020adversarial} & \checkmark & \textbf{94.1} \\ % Heatmaps and GANs
      \midrule
      Carreira \etal\cite{carreira2016human} & Partial${}^\dagger$ & 81.3 \\
      Sun \etal \cite{sun2017compositional} (Stage 0) & \xmark & 79.6  \\ 
      Sun \etal \cite{sun2017compositional} (Stage 1) & Partial${}^\dagger$  & 86.4 \\ % IEF (render prediction for another pass)
      Lathuili\`{e}re \etal \cite{lathuiliere2019comprehensive} & \xmark & 61.6  \\
        
      \midrule
      Ours & \xmark & 85.4 \\
      \bottomrule
    \end{tabular}
    %\label{tab:mpii_performance}
  \end{subtable}
  %\hfill\null
  \label{tab:sota_comparison}
\end{table}

\subsection{Ablation experiments}

\paragraph{Loss}
In this set of ablation experiments we compare our probabilistic $L_2$ multivariate Huber loss, equation (\ref{eq:loss}), to other probabilistic based losses. In particular we compare to losses using the negative log-likelihood (NLL) of the Gauss, Laplace and Charbonnier distributions:
\begin{align}
    \mathfrak{L}_{\text{\tiny Gauss}}(\by, \nu, A) &= -\log(|A|) + \|Ay-\nu\|_2^2/2\\
    \mathfrak{L}_{\text{\tiny Laplace}}(\by, \nu, A) &= -\log(|A|) + \|Ay-\nu\|_2\\
    \mathfrak{L}_{\text{\tiny Charb.}}(\by, \nu, A) &= -\log(|A|) + \sqrt{\|Ay-\nu\|_2^2+1}-1
\end{align}
We also compare the performance when we constrain the losses examined to have the identity, diagonal and full covariance matrices. Note when the covariance, $A$, is set to the identity matrix the above losses reduce to the mean squared error loss, mean error loss and the Charbonnier loss respectively. Our loss is reduced to the standard Huber loss when $A$ is set to be the identity matrix. The $\delta$ parameter was set to be the constant 1 for all Huber losses. Tuning this value might increase the performance for these methods. The results of these ablation experiments on WFLW are presented in tables \ref{tab:wflw_loss_ablation_NME} (standard dataset evaluation metric) and \ref{tab:wflw_loss_ablation_NLL} (NLL) and on MPII in table \ref{tab:mpii_loss_ablation_pckh} (standard dataset evaluation metric).

\begin{table}[htbp]
  \caption{WFLW NME}
  \centering
  \bobcaptionspace
  \begin{tabular}{l c c r}
        \toprule
        Distribution & Identity covariance ($\downarrow$) & Diagonal covariance ($\downarrow$) & Full covariance ($\downarrow$) \\
        \midrule
        Huber (ours)& 5.52 $\pm$ 0.02 & 4.91 $\pm$ 0.03 & 4.91 $\pm$ 0.04 \\
        Gauss       & 6.04 $\pm$ 0.10  & 5.65 $\pm$ 0.06 & 5.40  $\pm$ 0.17 \\
        Laplace     & 5.16 $\pm$ 0.03 & 4.90 $\pm$ 0.07 & 4.89 $\pm$ 0.02 \\
        Charbonnier & 5.64 $\pm$ 0.03 & 4.95 $\pm$ 0.02 & 4.96 $\pm$ 0.02 \\
        \bottomrule
    \end{tabular}
    \label{tab:wflw_loss_ablation_NME}
\end{table}
\begin{table}[htbp]
  \caption{\normalsize NLL WFLW}
  \centering
  \bobcaptionspace
    \begin{tabular}{l c r}
        \toprule
        Distribution & Diagonal covariance ($\downarrow$)& Full covariance ($\downarrow$) \\
        \midrule
        Huber (ours)& -307.0 $\pm$ 1.3 & -310.0 $\pm$ 1.4 \\
        Gauss       & -274.0 $\pm$ 2.1 & -279.3 $\pm$ 5.6 \\
        Laplace     & -306.6 $\pm$ 2.8 & -309.9 $\pm$ 1.2 \\
        Charbonnier & -305.9 $\pm$ 1.1 & -308.9 $\pm$ 0.8 \\
        \bottomrule
    \end{tabular}
  \label{tab:wflw_loss_ablation_NLL}
\end{table}

\begin{table}[htbp]
  \caption{\normalsize MPII PCKh@0.5}
    \centering
    \bobcaptionspace
    \begin{tabular}{l c c r}
        \toprule
        Distribution & Identity covariance ($\uparrow$) & Diagonal covariance ($\uparrow$) & Full covariance ($\uparrow$) \\
        \midrule
        Huber (ours)& 81.9 $\pm$ 0.1 & 85.1 $\pm$ 0.1 & 85.0 $\pm$ 0.1 \\
        Gauss       & 77.0 $\pm$ 0.1 & 81.3 $\pm$ 0.6 & 81.2 $\pm$ 0.5 \\
        Laplace     & 82.8 $\pm$ 0.1 & 85.0 $\pm$ 0.1 & 85.0 $\pm$ 0.1 \\
        Charbonnier & 80.2 $\pm$ 0.1 & 84.7 $\pm$ 0.1 & 84.7 $\pm$ 0.1 \\
        \bottomrule
    \end{tabular}
    \label{tab:mpii_loss_ablation_pckh}
\end{table}

\begin{comment}
\begin{table}[htbp]
  \caption{\normalsize MPII NLL}
  \centering
  \bobcaptionspace
    \begin{tabular}{l c r}
        \toprule
        Distribution & Diagonal covariance ($\downarrow$)& Full covariance ($\downarrow$) \\
        \midrule
        Huber (ours)& - $\pm$ - & - $\pm$ - \\
        Gauss       & - $\pm$ - & - $\pm$ - \\
        Laplace     & - $\pm$ - & - $\pm$ - \\
        Charbonnier & - $\pm$ - & - $\pm$ - \\
        \bottomrule
  \end{tabular}
  \hspace{0.05\textwidth}
  \label{tab:mpii_loss_ablation_NLL}
\end{table}
\end{comment}

%We evaluate the methods based on the NLL on the evaluation set as well as the standard metrics associated with the datasets, for reference see section \ref{sec:eval_metrics}.

%We also do ablations where we compare the performance when we constrain our method to estimate diagonal covariance matrix, compared to when allowing the covariance matrix to be any positive definite matrix.

By comparing column 1 with column 2 in table \ref{tab:wflw_loss_ablation_NME} we see that estimating the covariance matrix jointly with the position improves the performance of the position estimate. By comparing column 1 with column 2 in table \ref{tab:wflw_loss_ablation_NLL} we see that modeling the full covariance matrix, compared to modeling a diagonal matrix, consistently improves the NLL by approximately 3 units. The performance of the position estimate does not change significantly when modeling a full covariance matrix compared to only modeling a diagonal matrix, as can be seen by comparing column 2 with column 3 in table \ref{tab:wflw_loss_ablation_NME}.

From tables \ref{tab:wflw_loss_ablation_NLL} and \ref{tab:wflw_loss_ablation_NME} it is apparent that the Gauss loss has significantly worse performance compared to the other losses. This could be due to the fact that the tails for a Gauss distribution decays with $\mathcal{O}(e^{-r^2})$, whereas the other distributions have tails decaying with $\mathcal{O}(e^{-|r|})$. A distribution with quickly decaying tails tends to be less robust to outliers.

%The $\delta$ parameter was set to be the constant 1 for all Huber losses. Tuning this value might increase the performance for these methods.

% \input{figsAndTables/mpii_wflw_tta_ablations}

\paragraph{Other ablation experiments}

The supplementary material contains the results of more ablation studies, see section \ref{sec:supp_extra_tables}, w.r.t. the network architecture, the effect of replacing the final global pooling layer of the ResNet with trainable convolutions and the effect of training from a random initialization versus pre-training on large image repositories. A summary of the results 
are: both ResNet50 and ResNet101 produce better results than ResNet18, replacing the final global pooling with a convolution improves performance, pre-training improves performance and estimating the distributions mean indirectly by using the $\nu$ parameterization has similar performance to estimating the mean directly with the $\mu$ paramteterization, as described in section \ref{sec:method_multivar_huber}.

\subsection{Accuracy of probability estimates}

\begin{figure}[htbp]
\includegraphics[scale=0.4]{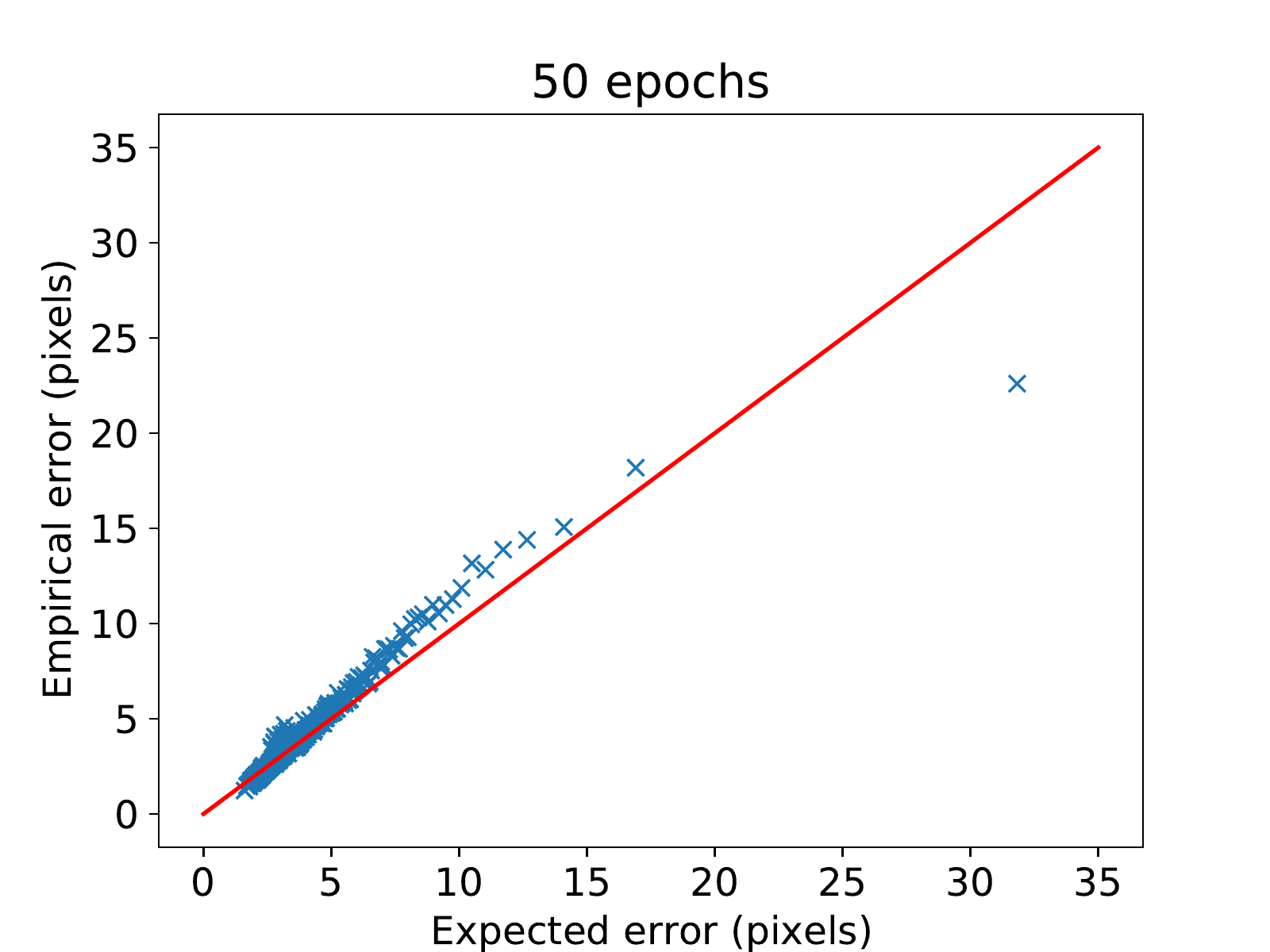}
\includegraphics[scale=0.4]{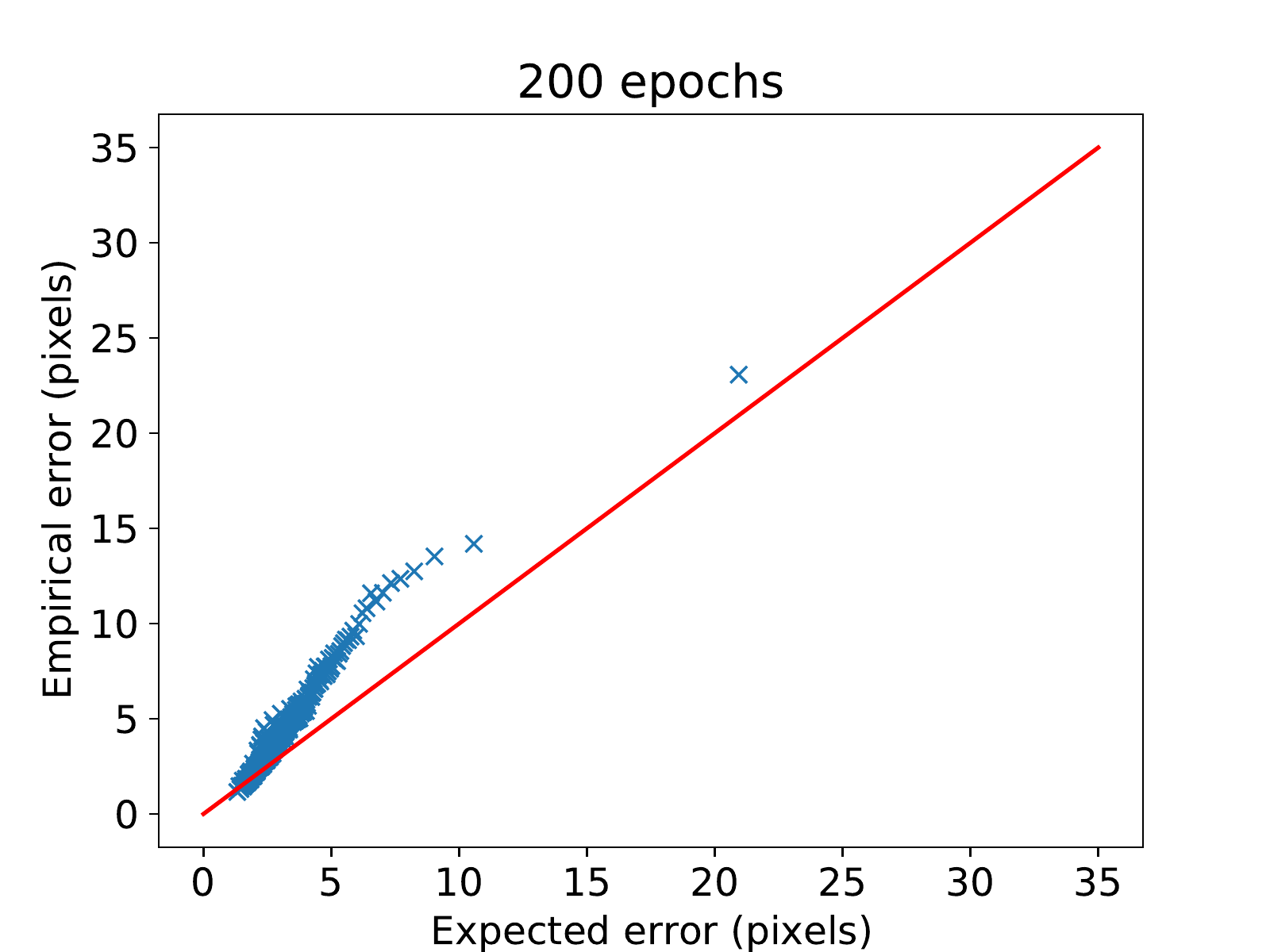} 
\bobcaptionspace
\caption{Relationship between empirical errors and expected errors based on the estimated precision matrix. 
Blue crosses correspond to the average empirical error compared to the average estimated error based on the variance of the output. The red line is the identity function. For a well calibrated method the blue crosses should be close to the red line.} 

\label{fig:scatter_emp_vs_est}
\end{figure}

Figure \ref{fig:scatter_emp_vs_est} displays plots of the expected error predicted for the test data in WFLW plotted against the actual average error as described in section \ref{sec:eval_metric_uncert_and_error}. We see that the network which is trained for 50 epochs predicts covariances which are well aligned with the empirical error, on average. The network which is trained for 200 epochs consistently underestimate the variance. In the supplementary material \ref{sec:supp_extra_tables} we can see that the 200 epochs network gets better NME performance but worse NLL performance. These observations are consistent with prior work \cite{guo2017calibration, NEURIPS2020_33cc2b87} which shows that assigned probabilities generally start to overfit earlier than the point estimate.

\begin{figure}[t]
\def\pwid{.3\textwidth}
\hspace{0.1\textwidth}
\includegraphics[width=\pwid,trim=100 20 100 20, clip]{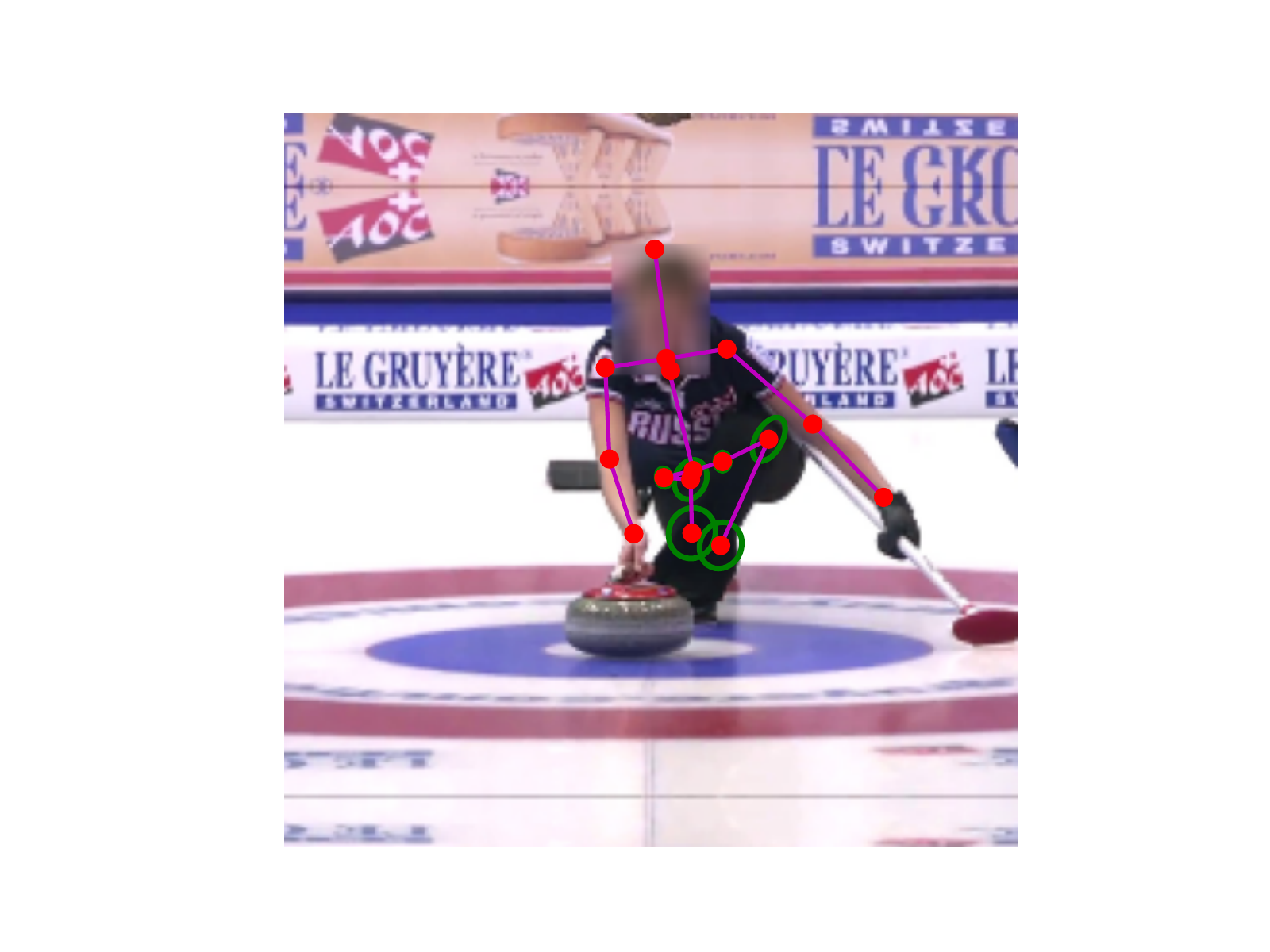}
\hspace{0.1\textwidth}
\includegraphics[width=\pwid,trim=100 20 100 20, clip]{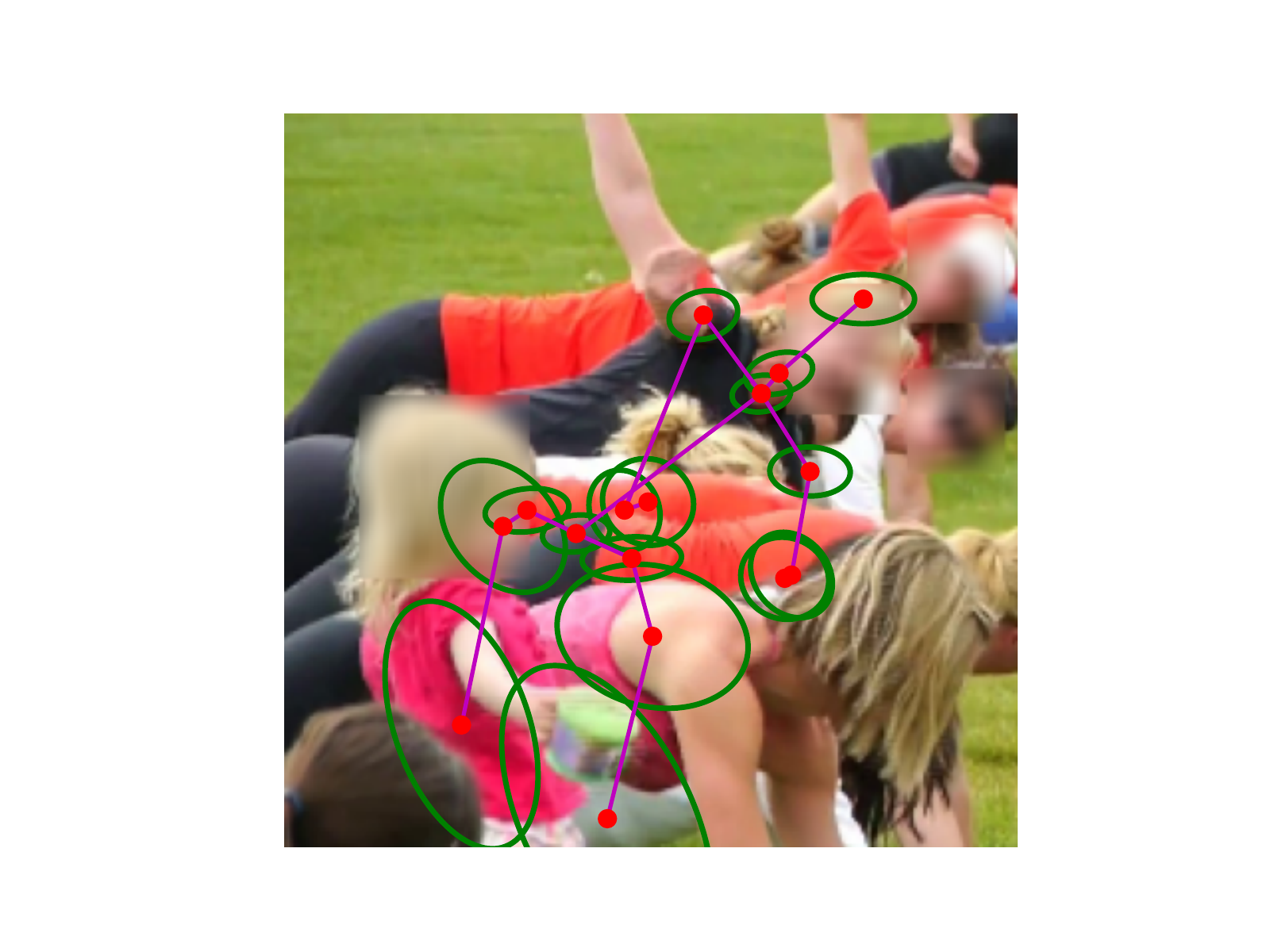} 
\caption{Qualitative illustration of results on MPII.
We show the predictions of our method. For each joint we show the mean position estimate (red) and corresponding uncertainties (the green ellipses correspond to one standard deviation). The test images above are specifically chosen to highlight the output uncertainties and how they correspond to prediction quality.
}
\label{fig:qual_errors}
\end{figure}

Qualitatively uncertainties tend to correspond to occlusion, strange poses or other ambiguities. In figure \ref{fig:qual_errors} we show two examples. In the left image most body parts are clearly visible and estimated uncertainties are low with low errors to ground truth locations. In the right image it is ambiguous which person to estimate the joint locations for, in addition to this most persons are also occluded. For the right image the errors are generally large with corresponding large estimated uncertainties.

\section{Conclusion \& Future work}
In this paper we have presented a way to enforce the output of neural networks to be positive definite matrices.
We have used this mapping to parameterize a unimodal probability distribution over $\mathbb{R}^N$.
Furthermore we show that the NLL loss with respect to this parameterization has bounded gradients among other desirable properties.

We have evaluated this method on standard face and body keypoint regression datasets and conclude that our method outperform other pure regression methods, however, the state of the art methods for this type of data still remain heatmap based.

One potential direction of research could be to use the network to estimate $\delta$ parameter of the Huber distribution.
Another direction could be to evaluate this method on higher dimensional data, such as regressing over 3d positions, an ubiquitous problem for real world applications.

\bibliography{refs}

\begin{thebibliography}{40}
\providecommand{\natexlab}[1]{#1}
\providecommand{\url}[1]{\texttt{#1}}
\expandafter\ifx\csname urlstyle\endcsname\relax
  \providecommand{\doi}[1]{doi: #1}\else
  \providecommand{\doi}{doi: \begingroup \urlstyle{rm}\Url}\fi

\bibitem[spe()]{spectrum_of_symm_real_mult}
Spectrum of symmetrizable matrix.
\newblock
  \url{https://math.stackexchange.com/questions/578891/spectrum-of-symmetrizable-matrix}.
\newblock Accessed: 2020-05-21.

\bibitem[Andriluka et~al.(2014)Andriluka, Pishchulin, Gehler, and
  Schiele]{andriluka14cvpr}
Mykhaylo Andriluka, Leonid Pishchulin, Peter Gehler, and Bernt Schiele.
\newblock 2d human pose estimation: New benchmark and state of the art
  analysis.
\newblock In \emph{Proceedings of the Conference on Computer Vision and Pattern
  Recognition (CVPR)}, 2014.

\bibitem[Aravkin(2010)]{aravkin2010robust}
Aleksandr Aravkin.
\newblock \emph{Robust methods for Kalman filtering/smoothing and bundle
  adjustment}.
\newblock PhD thesis, University of Washington, 2010.

\bibitem[{Barron}(2019)]{barron19cvpr}
J.~T. {Barron}.
\newblock A general and adaptive robust loss function.
\newblock In \emph{Proceedings of the Conference on Computer Vision and Pattern
  Recognition (CVPR)}, 2019.

\bibitem[Bin et~al.(2020)Bin, Cao, Chen, Ge, Tai, Wang, Li, Huang, Gao, and
  Sang]{bin2020adversarial}
Yanrui Bin, Xuan Cao, Xinya Chen, Yanhao Ge, Ying Tai, Chengjie Wang, Jilin Li,
  Feiyue Huang, Changxin Gao, and Nong Sang.
\newblock Adversarial semantic data augmentation for human pose estimation.
\newblock In \emph{Proceedings of the European Conference on Computer Vision
  (ECCV)}, 2020.

\bibitem[Bishop(2006)]{bishop2006pattern}
Christopher~M Bishop.
\newblock \emph{Pattern Recognition and Machine Learning}.
\newblock Springer, 2006.

\bibitem[Carreira et~al.(2016)Carreira, Agrawal, Fragkiadaki, and
  Malik]{carreira2016human}
Joao Carreira, Pulkit Agrawal, Katerina Fragkiadaki, and Jitendra Malik.
\newblock Human pose estimation with iterative error feedback.
\newblock In \emph{Proceedings of the Conference on Computer Vision and Pattern
  Recognition (CVPR)}, 2016.

\bibitem[Chen et~al.(2016)Chen, Kundu, Zhang, Ma, Fidler, and
  Urtasun]{Chen_2016_CVPR}
Xiaozhi Chen, Kaustav Kundu, Ziyu Zhang, Huimin Ma, Sanja Fidler, and Raquel
  Urtasun.
\newblock Monocular 3d object detection for autonomous driving.
\newblock In \emph{Proceedings of the Conference on Computer Vision and Pattern
  Recognition (CVPR)}, 2016.

\bibitem[Dapogny et~al.(2019)Dapogny, Bailly, and Cord]{dapogny2019decafa}
Arnaud Dapogny, Kevin Bailly, and Matthieu Cord.
\newblock De{C}a{FA}: deep convolutional cascade for face alignment in the
  wild.
\newblock In \emph{Proceedings of the Conference on Computer Vision and Pattern
  Recognition (CVPR)}, 2019.

\bibitem[Felzenszwalb and
  Huttenlocher(2005)]{Felzenszwalb03pictorialstructures}
Pedro~F. Felzenszwalb and Daniel~P. Huttenlocher.
\newblock Pictorial structures for object recognition.
\newblock \emph{International Journal of Computer Vision (IJCV)}, 61\penalty0
  (1):\penalty0 55--79, 2005.

\bibitem[Feng et~al.(2018)Feng, Kittler, Awais, Huber, and Wu]{Feng_2018_CVPR}
Zhen-Hua Feng, Josef Kittler, Muhammad Awais, Patrik Huber, and Xiao-Jun Wu.
\newblock Wing loss for robust facial landmark localisation with convolutional
  neural networks.
\newblock In \emph{Proceedings of the Conference on Computer Vision and Pattern
  Recognition (CVPR)}, 2018.

\bibitem[Girshick(2015)]{Girshick_2015_ICCV}
Ross Girshick.
\newblock Fast r-cnn.
\newblock In \emph{Proceedings of the International Conference on Computer
  Vision (ICCV)}, December 2015.

\bibitem[Gundavarapu et~al.(2019)Gundavarapu, Srivastava, Mitra, Sharma, and
  Jain]{Gundavarapu_2019_CVPR_Workshops}
Nitesh~B. Gundavarapu, Divyansh Srivastava, Rahul Mitra, Abhishek Sharma, and
  Arjun Jain.
\newblock Structured aleatoric uncertainty in human pose estimation.
\newblock In \emph{Proceedings of the Conference on Computer Vision and Pattern
  Recognition Workshops (CVPRW)}, 2019.

\bibitem[Guo et~al.(2017)Guo, Pleiss, Sun, and Weinberger]{guo2017calibration}
Chuan Guo, Geoff Pleiss, Yu~Sun, and Kilian~Q Weinberger.
\newblock On calibration of modern neural networks.
\newblock In \emph{International Conference on Machine Learning}, pages
  1321--1330. PMLR, 2017.

\bibitem[He et~al.(2016)He, Zhang, Ren, and Sun]{he2016deep}
Kaiming He, Xiangyu Zhang, Shaoqing Ren, and Jian Sun.
\newblock Deep residual learning for image recognition.
\newblock In \emph{Proceedings of the IEEE conference on computer vision and
  pattern recognition}, pages 770--778, 2016.

\bibitem[Ionescu et~al.(2015)Ionescu, Vantzos, and
  Sminchisescu]{ionescu2015matrix}
Catalin Ionescu, Orestis Vantzos, and Cristian Sminchisescu.
\newblock Matrix backpropagation for deep networks with structured layers.
\newblock In \emph{Proceedings of the Conference on Computer Vision and Pattern
  Recognition (CVPR)}, 2015.

\bibitem[Kuhn(1973)]{kuhn1973note}
Harold~W Kuhn.
\newblock A note on fermat's problem.
\newblock \emph{Mathematical programming}, 4\penalty0 (1):\penalty0 98--107,
  1973.

\bibitem[Kumar et~al.(2020)Kumar, Marks, Mou, Wang, Jones, Cherian,
  Koike-Akino, Liu, and Feng]{kumar2020luvli}
Abhinav Kumar, Tim~K Marks, Wenxuan Mou, Ye~Wang, Michael Jones, Anoop Cherian,
  Toshiaki Koike-Akino, Xiaoming Liu, and Chen Feng.
\newblock {LUVLi} face alignment: Estimating landmarks' location, uncertainty,
  and visibility likelihood.
\newblock In \emph{Proceedings of the Conference on Computer Vision and Pattern
  Recognition (CVPR)}, 2020.

\bibitem[Lathuili{\`e}re et~al.(2019)Lathuili{\`e}re, Mesejo, Alameda-Pineda,
  and Horaud]{lathuiliere2019comprehensive}
St{\'e}phane Lathuili{\`e}re, Pablo Mesejo, Xavier Alameda-Pineda, and Radu
  Horaud.
\newblock A comprehensive analysis of deep regression.
\newblock \emph{IEEE Transactions on Pattern Analysis and Machine Intelligence
  (TPAMI)}, 42\penalty0 (9):\penalty0 2065--2081, 2019.

\bibitem[Liu et~al.(2018)Liu, Ok, Vega-Brown, and Roy]{Liu_2018_ICRA}
Katherine Liu, Kyel Ok, William Vega-Brown, and Nicholas Roy.
\newblock Deep inference for covariance estimation: Learning gaussian noise
  models for state estimation.
\newblock In \emph{IEEE International Conference on Robotics and Automation
  (ICRA)}, May 2018.

\bibitem[Mohlin et~al.(2020)Mohlin, Bianchi, and
  Sullivan]{NEURIPS2020_33cc2b87}
David Mohlin, G\'{e}rald Bianchi, and Josephine Sullivan.
\newblock {Probabilistic Orientation Estimation with Matrix Fisher
  Distributions}.
\newblock In \emph{Advances in Neural Information Processing Systems
  (NeurIPs)}, 2020.

\bibitem[Mousavian et~al.(2017)Mousavian, Anguelov, Flynn, and
  Kosecka]{Mousavian_2017_CVPR}
Arsalan Mousavian, Dragomir Anguelov, John Flynn, and Jana Kosecka.
\newblock 3d bounding box estimation using deep learning and geometry.
\newblock In \emph{Proceedings of the Conference on Computer Vision and Pattern
  Recognition (CVPR)}, 2017.

\bibitem[Newell et~al.(2016)Newell, Yang, and
  Deng]{newell16eccv_stackedhourglass}
Alejandro Newell, Kaiyu Yang, and Jia Deng.
\newblock Stacked hourglass networks for human pose estimation.
\newblock In \emph{Proceedings of the European Conference on Computer Vision
  (ECCV)}, 2016.

\bibitem[Qian et~al.(2019)Qian, Sun, Wu, Qian, and Jia]{qian2019aggregation}
Shengju Qian, Keqiang Sun, Wayne Wu, Chen Qian, and Jiaya Jia.
\newblock Aggregation via separation: Boosting facial landmark detector with
  semi-supervised style translation.
\newblock In \emph{Proceedings of the Conference on Computer Vision and Pattern
  Recognition (CVPR)}, 2019.

\bibitem[Redmon and Farhadi(2018)]{yolov3}
Joseph Redmon and Ali Farhadi.
\newblock Yolov3: An incremental improvement.
\newblock \emph{arXiv}, 2018.

\bibitem[Sarafianos et~al.(2016)Sarafianos, Boteanu, Ionescu, and
  Kakadiaris]{Sarafianos_2016_CVIU}
Nikolaos Sarafianos, Bogdan Boteanu, Bogdan Ionescu, and Ioannis~A. Kakadiaris.
\newblock 3d human pose estimation: A review of the literature and analysis of
  covariates.
\newblock \emph{Computer Vision and Image Understanding (CVIU)}, 2016.

\bibitem[Su et~al.(2019)Su, Ye, Zhang, Dai, and Sheng]{su2019cascade}
Zhihui Su, Ming Ye, Guohui Zhang, Lei Dai, and Jianda Sheng.
\newblock Cascade feature aggregation for human pose estimation.
\newblock \emph{arXiv preprint arXiv:1902.07837}, 2019.

\bibitem[Sun et~al.(2019)Sun, Xiao, Liu, and Wang]{Sun_2019_CVPR}
Ke~Sun, Bin Xiao, Dong Liu, and Jingdong Wang.
\newblock Deep high-resolution representation learning for human pose
  estimation.
\newblock In \emph{Proceedings of the Conference on Computer Vision and Pattern
  Recognition (CVPR)}, 2019.

\bibitem[Sun et~al.(2017)Sun, Shang, Liang, and Wei]{sun2017compositional}
Xiao Sun, Jiaxiang Shang, Shuang Liang, and Yichen Wei.
\newblock Compositional human pose regression.
\newblock In \emph{Proceedings of the Conference on Computer Vision and Pattern
  Recognition (CVPR)}, 2017.

\bibitem[Tang et~al.(2018)Tang, Yu, and Wu]{Tang_2018_ECCV}
Wei Tang, Pei Yu, and Ying Wu.
\newblock Deeply learned compositional models for human pose estimation.
\newblock In \emph{Proceedings of the European Conference on Computer Vision
  (ECCV)}, 2018.

\bibitem[Tang et~al.(2020)Tang, Peng, Li, and Metaxas]{tang20_dunet}
Zhiqiang Tang, Xi~Peng, Kang Li, and Dimitris~N. Metaxas.
\newblock Towards efficient {U-Nets}: A coupled and quantized approach.
\newblock \emph{IEEE Transactions on Pattern Analysis and Machine Intelligence
  (TPAMI)}, 42\penalty0 (8):\penalty0 2038--2050, 2020.

\bibitem[Tompson et~al.(2014)Tompson, Jain, LeCun, and Bregler]{tompsonjoint}
Jonathan Tompson, Arjun Jain, Yann LeCun, and Christoph Bregler.
\newblock Joint training of a convolutional network and a graphical model for
  human pose estimation.
\newblock In \emph{Advances in Neural Information Processing Systems
  (NeurIPs)}, 2014.

\bibitem[Tompson et~al.(2015)Tompson, Goroshin, Jain, LeCun, and
  Bregler]{tompson2015efficient}
Jonathan Tompson, Ross Goroshin, Arjun Jain, Yann LeCun, and Christoph Bregler.
\newblock Efficient object localization using convolutional networks.
\newblock In \emph{Proceedings of the Conference on Computer Vision and Pattern
  Recognition (CVPR)}, 2015.

\bibitem[Toshev and Szegedy(2014)]{Toshev_2014_CVPR}
Alexander Toshev and Christian Szegedy.
\newblock Deeppose: Human pose estimation via deep neural networks.
\newblock In \emph{Proceedings of the Conference on Computer Vision and Pattern
  Recognition (CVPR)}, June 2014.

\bibitem[Wang et~al.(2019)Wang, Bo, and Fuxin]{Wang_2019_ICCV}
Xinyao Wang, Liefeng Bo, and Li~Fuxin.
\newblock Adaptive wing loss for robust face alignment via heatmap regression.
\newblock In \emph{Proceedings of the International Conference on Computer
  Vision (ICCV)}, 2019.

\bibitem[Wu et~al.(2018)Wu, Qian, Yang, Wang, Cai, and Zhou]{wayne2018lab}
Wayne Wu, Chen Qian, Shuo Yang, Quan Wang, Yici Cai, and Qiang Zhou.
\newblock Look at boundary: A boundary-aware face alignment algorithm.
\newblock In \emph{CVPR}, June 2018.

\bibitem[Wu and Yang(2017)]{wu2017leveraging}
Wenyan Wu and Shuo Yang.
\newblock Leveraging intra and inter-dataset variations for robust face
  alignment.
\newblock In \emph{Proceedings of the Conference on Computer Vision and Pattern
  Recognition Workshops (CVPRW)}, 2017.

\bibitem[Yang et~al.(2017)Yang, Li, Ouyang, Li, and Wang]{yang2017learning}
Wei Yang, Shuang Li, Wanli Ouyang, Hongsheng Li, and Xiaogang Wang.
\newblock Learning feature pyramids for human pose estimation.
\newblock In \emph{Proceedings of the International Conference on Computer
  Vision (ICCV)}, 2017.

\bibitem[Zhu et~al.(2015)Zhu, Li, Change~Loy, and Tang]{zhu2015face}
Shizhan Zhu, Cheng Li, Chen Change~Loy, and Xiaoou Tang.
\newblock Face alignment by coarse-to-fine shape searching.
\newblock In \emph{Proceedings of the Conference on Computer Vision and Pattern
  Recognition (CVPR)}, 2015.

\bibitem[Zhu et~al.(2016)Zhu, Lei, Liu, Shi, and Li]{zhu2016face}
Xiangyu Zhu, Zhen Lei, Xiaoming Liu, Hailin Shi, and Stan~Z Li.
\newblock Face alignment across large poses: A 3d solution.
\newblock In \emph{Proceedings of the Conference on Computer Vision and Pattern
  Recognition (CVPR)}, 2016.

\end{thebibliography}
\newpage
\appendix

\section{Properties of Huber distribution}
\subsection{Normalizing factor of Huber distribution}
\label{sec:supp_normalizing_factor_and_var}

Recall

\begin{equation}
  p(\mathbf{x} \mid \nu, \Lambda, \delta) \propto \exp\left(-h_{\delta}\left(\|\Lambda^{1/2}(\mathbf{x}-\nu)\|_2\right)\right)
\end{equation}

To make this integrate to 1 we need a normalizing factor $Z$. It will be a function of the parameters of the distribution, i.e. $\nu$, $\Lambda$ and $\delta$. First we consider how $\nu$ and $\Lambda$ influence the normalizing factor. By doing the variable substitution $\mathbf{y} = \Lambda^{1/2}(\mathbf{x}-\nu)$ we get the following

\begin{align}
    1 &= \dfrac {1} {Z(\Lambda, \nu, \delta)} \int\limits_{\mathbf{x} \in \mathbb{R}^d} p(\mathbf{x} \mid \mu, \Lambda, \delta) d\mathbf{x} \\
    &= \dfrac {1} {Z(\Lambda, \nu, \delta)|\Lambda^{1/2}|} \int\limits_{\mathbf{y} \in \mathbb{R}^d} p(\mathbf{y} \mid 0, I, \delta) d\mathbf{y} \\
    &= \dfrac {c_d(\delta)} {Z(\Lambda, \nu, \delta)|\Lambda^{1/2}|}
\end{align}

Solving this with respect to Z gives

\begin{equation}
    Z(\Lambda, \nu, \delta) = \dfrac {c_d(\delta)} {|\Lambda^{1/2}|}
\end{equation}

By using the normalizing factor we get the pdf for the distribution

\begin{equation}
  p(\mathbf{x} \mid \nu, \Lambda, \delta) = \dfrac{|\Lambda|^{1/2}} {c_d(\delta)} \exp\left(-h_{\delta}\left(\|\Lambda^{1/2}(\mathbf{x}-\nu)\|_2\right)\right)
\end{equation}

We find the expression of $c_d(\delta)$ by evaluating the integral which defines it. By doing a change to spherical coordinates and using radial symmetry we get

\begin{align}
    {c_d(\delta)} &= \int\limits_{0}^{\infty} |S_{d-1}| r^{d-1} exp(-h_{\delta}(r)) dr \\
    &= |S_{d-1}| (\int\limits_{0}^{\delta}  r^{d-1} exp(-r^2/2) dr + \int\limits_{\delta}^{\infty}  r^{d-1} exp(-\delta r + \delta^2/2) dr) \\
     &= |S_{d-1}| (a(d-1, \delta) + exp(\delta^2/2) b(d-1, \delta))
\end{align}

Where $|S_d|$ is the volume of a d dimensional unit sphere $S_d = \{x \in \mathbb{R}^{d+1}: \text{ } ||x||_2 = 1\}$.
a and b are defined by the two integrals.

We first notice 
\begin{equation} 
a(0, \delta) = \int\limits_{0}^\delta exp(-r^2/2) dr = \sqrt{\dfrac {\pi} {2}}erf(\delta/\sqrt{2})
\end{equation}
and
\begin{equation}
    a(1, \delta) = \int\limits_{0}^\delta r* exp(-r^2/2) dr = \left[ -exp(-r^2/2) \right]_{0}^{\delta} = 1-exp(\delta^2/2)
\end{equation}

by performing integration by parts we get 

\begin{align}
    a(n, \delta) &= 
    \int\limits_{0}^\delta r^{n-1} r* exp(-r^2/2) dr \\
    &= \left[ -r^{n-1} exp(-r^2/2) \right]_{0}^{\delta} + \int\limits_{0}^{\delta} (n-1)r^{n-2} exp(-r^2/2) \\
    &= - \delta^{n-1} exp(-\delta^2/2) + (n-1)a(n-2, \delta)
\end{align}

from this we have recursively defined $a(n, \delta)$ for all values.

similarly for b
\begin{equation}
    b(0, \delta) = 
    \int\limits_{\delta}^\infty exp(-\delta r) dr =
    \dfrac {exp(-\delta^2)} {\delta}
\end{equation}

\begin{align}
    b(n, \delta) &= 
    \int\limits_{\delta}^\infty r^{n} exp(-\delta r) dr \\
    &= \left[ -\dfrac{r^{n}} {\delta} exp(-\delta r) \right]_{\delta}^{\infty} + \int\limits_{\delta}^{\infty} n r^{n-1} exp(-\delta r) dr \\
    &= \delta^{n-1} exp(-\delta^2) + nb(n-1, \delta)
\end{align}

we have now defined a and b for all  $n$ and $\delta$, thereby also the normalizing the normalizing constant for all $d$ and $\delta$.

\subsection{Variance of huber distribution}
\label{sec:supp_variance_of_huber}

The variance for $p(\mathbf{x} \mid 0, I, \delta)$ can be found by 

\begin{equation}
    E(||\mathbf{X}||_2^2) =
    \dfrac {|S_{d-1}|}
           {c_d(\delta)}
    \int\limits_0^{\infty} r^{d+1}exp(-h_{\delta}(r))dr = \\
    \dfrac {a(d+1, \delta) + exp(\delta^2/2)b(d+1, \delta)}
           {a(d-1, \delta) + exp(\delta^2/2) b(d-1, \delta)}
\end{equation}

Due to symmetry we know that $Var(\mathbf{X})$ is a diagonal matrix and $tr(Var(\mathbf{X})) = E(||\mathbf{X}||_2^2)$
therefore

\begin{equation}
    Var(\mathbf{X}) = \dfrac {a(d+1, \delta) + exp(\delta^2/2)b(d+1, \delta)}
           {d(a(d-1, \delta) + exp(\delta^2/2) b(d-1, \delta))}
\end{equation}

We get the following after doing a variable substitution $\by = \Sigma^{1/2} \bx$

\begin{equation}
    Var(\mathbf{Y}) = \dfrac {a(d+1, \delta) + exp(\delta^2/2)b(d+1, \delta)}
           {d(a(d-1, \delta) + exp(\delta^2/2) b(d-1, \delta))} \Sigma
\end{equation}

The expected distance between the mean and a sample will then be
\begin{gather}
    E(||\mathbf{Y}||_2^2) = E(tr(\mathbf{Y}^T\mathbf{Y})) = E(tr(\mathbf{Y}\mathbf{Y}^T)) = \\ \dfrac {a(d+1, \delta) + exp(\delta^2/2)b(d+1, \delta)}
           {d(a(d-1, \delta) + exp(\delta^2/2) b(d-1, \delta))} tr(\Sigma)
\end{gather}

Specifically for d = 2, $\delta=1$

\begin{gather}
    E(||\mathbf{Y}||_2^2) \approx 3.07 tr(\Sigma)
    \label{eq:expected_error}
\end{gather}

\section{Equation for gradients when applying function on eigenvalues} \label{sec:sup_mat_diag_remapping}
If we have the square symmetric matrix B with eigendecomposition $B =V^TDV$ and define 
$A = G(B) = V^T diag(g(D_{1,1}), g(D_{2,2}), \cdots g(D_{d,d}))V$ where $g$ is a differentiable function $g: \mathbb{R} \rightarrow \mathbb{R}$. Then the gradient of a function $L$ can be computed with respect to B through the following equation.
\begin{equation}
\dfrac{\partial L} {\partial B} = V^T(V (\dfrac {\partial L} {\partial A}) V^T \circ K(D, g))V
\label{eq:diag_remapping_theorem}
\end{equation}

where
\begin{gather}
    K(D, g)_{i,j} = \begin{cases}
    \dfrac{g(D)_{i,i} - g(D)_{j,j}} {D_{i,i}-D_{j,j}} \text{ if } D_{i,i} \ne D_{j,j} \\
    g'(D_{i,i}) \text{ otherwise}    \end{cases}
\end{gather}

and $\circ$ is a elementwise multiplication. This expression is similar to the expressions in \cite{ionescu2015matrix}, except it handles the case when different eigenvalues are equal as well.

Note that $\dfrac {\partial L} {\partial A}$ and $\dfrac {\partial L} {\partial B}$ needs to be symmetric matrices since A and B are symmetric.

\subsection{Proof}
\subsubsection{Reduce proof to diagonal matrices}
given a matrix $B = V^T D V$ pick the constant $\hat{V} = V$, note V is a variable dependent on B while $\hat{V}$ is constant.

Define $C = \hat{V} A \hat{V}^T \implies \hat{V}^T C \hat{V}= A$ and $E = \hat{V} B \hat{V}^T \implies \hat{V}^T E \hat{V} = B$

First
\begin{equation}
    \dfrac {\partial L} {\partial C_{i,j}} =  \sum\limits_{m=1}^d\sum\limits_{n=1}^d\hat{V}_{i,m} \dfrac {\partial L} {\partial A_{m,n}} (\hat{V}^T)_{n,j} = 
    (\hat{V} \dfrac {\partial L} {\partial A} \hat{V}^T)_{i,j}
    \label{eq:basis_change_diff_1}
\end{equation}

The same holds for any multiplication of constant matrices.

Such as 

\begin{equation}
    \dfrac {\partial L} {\partial B_{i,j}} = 
    (\hat{V}^T \dfrac {\partial L} {\partial E} \hat{V})_{i,j}
    \label{eq:basis_change_diff_2}
\end{equation}

Since C and E are diagonal this further simplifies our proof.

\subsubsection{Differentiation of diagonal elements}

We will use the single entry matrix $J^{i,j}$ in following sections. The dimension of this matrix is implicit based on context.
\begin{equation}
J^{i,j}_{m,n} = \mathbb{1}(i=m \land j=n)
\end{equation}
where $\mathbb{1}$ is the indicator function.

If E is diagonal then 
$F = E+J^{i,i}\epsilon$ is trivially diagonal as well,
therefore
\begin{equation}
\lim_{\epsilon \rightarrow 0} \dfrac {G(E)-G(F)} {\epsilon} = J^{i,i} g'(E_{i,i}) \forall i
\end{equation}

Since $C = G(E)$ we get

\begin{equation}
    \dfrac {\partial C}
    {\partial E_{i,i}} = J^{i,i}g'(E_{i,i})
    \label{eq:diag_proof_diag_elements}
\end{equation}

\subsubsection{Differentiation of non-diagonal elements}

Let's consider how $g(E)$ changes when we change the element of row i and column j. Since E is diagonal this will only affect the i:th and j:th eigenvalues and eigenvectors. Without loss of generality we can analyze the case when we change the non-diagonal elements of a $2\times 2$ matrix.

\begin{equation}
E = \begin{bmatrix}
x & 0 \\
0 & y
\end{bmatrix}
\end{equation}

First we analyze the case when $x \ne y$

We can find the eigenvalues of $E + \epsilon(J^{1,2} + J^{2,1})$ by solving 
$|E + \epsilon(J^{1,2} + J^{2,1}) - \lambda I| = 0$ for $\lambda$

The solution of this is 
\begin{equation}
    \lambda = \dfrac {x+y} {2} \pm \sqrt{\left(\dfrac {x-y} {2}\right)^2+\epsilon^2} = \\
    \dfrac {x+y} {2} \pm \left(\dfrac {|x-y|} {2} +\dfrac {\epsilon^2} {|x-y|} + \mathcal{O}(\epsilon^3)\right)= \\
\end{equation} 

The first step can be done by completing the square and the second step is the first terms of the maclaurin series.

Assume x > y then solve for eigenvectors to get
\begin{equation}
\mathbf{v}^T (x - \dfrac {x+y} {2} - \dfrac {|x-y|} {2} - \dfrac {\epsilon^2} {|x-y|} + \mathcal{O}(\epsilon^3), \epsilon) = 
\mathbf{v}^T (-\dfrac {\epsilon^2} {|x-y|} + \mathcal{O}(\epsilon^3),\epsilon)
\end{equation}

Solving for $\mathbf{v}$ we get

\begin{equation}
  \mathbf{v} = \left[1, \dfrac {\epsilon} {|x-y|}  + \mathcal{O}(\epsilon^2) \right]
\end{equation}

The normalizing factor for $\mathbf{v}$ will be $1 + \mathcal{O}(\epsilon^2)$ so it will not influence the limit of the derivative.
If $y > x$ the sign of the epsilon term would change.

Our new basis is now

\begin{equation}
    \dfrac{\begin{bmatrix}
        1 & -\dfrac {\epsilon} {x-y} + \mathcal{O}(\epsilon^2) \\
        \dfrac {\epsilon} {x-y}+ \mathcal{O}(\epsilon^2) & 1
    \end{bmatrix}}
    {(1+\mathcal{O}(\epsilon^2))^2}
\end{equation}

Putting it together

\begin{align}
    &E + \epsilon(J^{1,2}+ J^{2,1})  \\
    = \dfrac {1} {(1+\mathcal{O}(\epsilon^2))^2}
    &\begin{bmatrix}
        1 & -\dfrac {\epsilon} {x-y}+\mathcal{O}(\epsilon^2) \\
        \dfrac {\epsilon} {x-y}+\mathcal{O}(\epsilon^2) & 1
    \end{bmatrix} \\ \times
    &\begin{bmatrix}
        x + \mathcal{O}(\epsilon^2) & 0 \\
        0 & y + \mathcal{O}(\epsilon^2)
    \end{bmatrix} \\
    \times
    &\begin{bmatrix}
        1 & \dfrac {\epsilon} {x-y}+\mathcal{O}(\epsilon^2) \\
        -\dfrac {\epsilon} {x-y}+\mathcal{O}(\epsilon^2) & 1
    \end{bmatrix}
\end{align}

Where $\times$ is the standard matrix multiplication.
Applying $g$ on the diagonal terms gives

\begin{align}
    &G(E + \epsilon(J^{1,2} + J^{2,1})) \\ =
    \dfrac {1} {(1+\mathcal{O}(\epsilon^2))^2} &\begin{bmatrix}
        1 & -\dfrac {\epsilon} {x-y}+\mathcal{O}(\epsilon^2) \\
        \dfrac {\epsilon} {x-y}+\mathcal{O}(\epsilon^2) & 1
    \end{bmatrix} \\ \times
    &\begin{bmatrix}
        g(x) + \mathcal{O}(\epsilon^2) & 0 \\
        0 & g(y) + \mathcal{O}(\epsilon^2)
    \end{bmatrix} \\ \times
    &\begin{bmatrix}
        1 & \dfrac {\epsilon} {x-y}+\mathcal{O}(\epsilon^2) \\
        -\dfrac {\epsilon} {x-y}+\mathcal{O}(\epsilon^2) & 1
    \end{bmatrix} \\ =
    \dfrac {1} {(1+\mathcal{O}(\epsilon^2))^2} &\begin{bmatrix}
        1 & -\dfrac {\epsilon} {x-y} + \mathcal{O}(\epsilon^2) \\
        \dfrac {\epsilon} {x-y}+ \mathcal{O}(\epsilon^2) & 1
    \end{bmatrix} \\ \times
    &\begin{bmatrix}
        g(x) + \mathcal{O}(\epsilon^2) & \epsilon \dfrac {g(x)} {x-y} + \mathcal{O}(\epsilon^3) \\
        -\epsilon \dfrac {g(y)} {x-y} + \mathcal{O}(\epsilon^3) & g(y) + \mathcal{O}(\epsilon^2)
    \end{bmatrix} \\ =
    \dfrac {1} {(1+\mathcal{O}(\epsilon^2))^2}
    &\begin{bmatrix}
        g(x) + \mathcal{O}(\epsilon^2) & \epsilon \dfrac {g(x)-g(y)} {x-y} + \mathcal{O}(\epsilon^3) \\
        \epsilon \dfrac {g(x)-g(y)} {x-y} + \mathcal{O}(\epsilon^3) & g(y) + \mathcal{O}(\epsilon^2)
    \end{bmatrix}
\end{align}

The first step comes from the fact that $g$ is continous. The other two steps are matrix multiplications.

From this it is obvious that 

\begin{equation}
    \dfrac {\partial {G(E)}} {\partial E_{1,2}} = \dfrac {g(x)-g(y)} {x-y} (J^{1,2} + J^{2,1})
    \label{eq:diag_proof_x_is_not_y}
\end{equation}

Note $E_{1,2} = E_{2,1}$ since E is symmetric.

\paragraph{Differentiation of non-diagonal when x=y}
We do the same procedure and solve the eigenvalues to be
\begin{equation}
    \lambda = x \pm \epsilon
\end{equation}

We solve for eigenvectors and get

\begin{equation}
    \mathbf{v}^T \left[x - x - \epsilon, \epsilon \right] = 0
\end{equation}
which gives
\begin{equation}
    \mathbf{v} = \left[\dfrac{1} {\sqrt{2}}, \dfrac{1} {\sqrt{2}}\right]
\end{equation}

Therefore
\begin{gather}
    \begin{bmatrix}
         x &\epsilon \\
         \epsilon & x
    \end{bmatrix} = 
    \begin{bmatrix}
         1/\sqrt{2} & -1/\sqrt{2} \\
         1/\sqrt{2} & 1/\sqrt{2}
    \end{bmatrix}
    \begin{bmatrix}
         x + \epsilon & 0 \\
         0 &x - \epsilon \\
    \end{bmatrix} = 
    \begin{bmatrix}
         1/\sqrt{2} & 1/\sqrt{2} \\
         -1/\sqrt{2} & 1/\sqrt{2}
    \end{bmatrix}
\end{gather}

This is also trivially verified by matrix multiplication.

\begin{align}
    &G(\begin{bmatrix}
         x &\epsilon \\
         \epsilon & x
    \end{bmatrix}) \\ = 
    &\begin{bmatrix}
         1/\sqrt{2} & -1/\sqrt{2} \\
         1/\sqrt{2} & 1/\sqrt{2}
    \end{bmatrix} \\
    \times&\begin{bmatrix}
         g(x) + \epsilon g'(x) + \mathcal{O}(\epsilon^2) & 0 \\
         0 & g(x) - \epsilon g'(x) +  \mathcal{O}(\epsilon^2) \\
    \end{bmatrix} \\
    \times&\begin{bmatrix}
         1/\sqrt{2} & 1/\sqrt{2} \\
         -1/\sqrt{2} & 1/\sqrt{2}
    \end{bmatrix} \\
    = &G(\begin{bmatrix}
         g(x) &\epsilon g'(x) \\
         \epsilon g'(x) & g(x)
    \end{bmatrix}) + \mathcal{O}(\epsilon^2)
\end{align}

From this we see that

\begin{equation}
    \dfrac {\partial {g(E)}} {\partial E_{1,2}} = g'(x) (J^{1,2} + J^{2,1})
    \label{eq:diag_proof_x_is_y}
\end{equation}

When $x=y$

\subsubsection{Wrapping up the proof}

From the earlier argument this will now hold for all square diagonal matrices

By combining equations \ref{eq:diag_proof_x_is_y}, \ref{eq:diag_proof_x_is_not_y} and \ref{eq:diag_proof_diag_elements} we get

\begin{equation}
    \dfrac{\partial L} {\partial E_{i,j}} =  \left(\dfrac{\partial L} {\partial C_{i,j}} \right)
    * \begin{cases} \dfrac {g(E_{i,i}) - g(E_{j,j})} {E_{i,i} - E_{j,j}} \text { if } E_{i,i} \ne E_{j,j} \\
    g'(E_{i,i}) \text { otherwise}
    \end{cases}
    \label{eq:diag_proof_almost_done}
\end{equation}

By combining equations \ref{eq:basis_change_diff_1}, \ref{eq:basis_change_diff_2} and \ref{eq:diag_proof_almost_done} we can construct a proof for equation \ref{eq:diag_remapping_theorem} for every matrix B.

\subsubsection{Final comments}
In practice we use the gradient when the two eigenvalues are sufficiently close instead of identical to avoid numerical instability.

\section{Loss} \label{sec:supp_loss}

In this section we prove that under the assumption that $||\bx||_2$ is bounded, our suggested loss has bounded gradients, is convex for the convex set when all eigenvalues of $A$ are larger than $\theta$. and has bounded Hessians for the same set.

From now on we will only analyze the case $\delta = 1$ since that is the value we use for all experiments.

Recall that our loss is parameterized as

\begin{equation}
    p(\bx \mid \nu, A) = \dfrac {|A|} {c_d(1)} \exp(-h_1(||A\bx-\nu||_2))
\end{equation}

The negative log likelihood of this function, denoted $\mathfrak{L}$ is:

\begin{equation}
    \mathfrak{L}(\bx, \nu, A) = -log(|A|) + log(c_d(1)) + h(||A\bx-\nu||_2)
\end{equation}

What remains is to show that $-log(|A|)$ and $h(||A\bx-\nu||_2)$ have these properties with respect to $A$ and $\nu$. Note this is stronger than convex with respect to the two variables individually.
Since we need
\begin{equation}
    \mathfrak{L}(\bx, \lambda \nu_1 + (1-\lambda) \nu_2, \lambda A_1 + (1-\lambda) A_2) \le \lambda \mathfrak{L}(\bx, \nu_1, A_1) + (1-\lambda) \mathfrak{L}(\bx, \nu_2, A_2)
\end{equation}

\subsection{Study of diagonal remapping function}
This section is for future reference in the proof.
Recall that the function we apply on eigenvalues is 
\begin{equation}
g(\lambda) = \begin{cases}
    \lambda & \text{if $\lambda > \theta$} \\
    \theta \exp\left(\lambda/\theta - 1 \right) & \text{otherwise}
    \end{cases}
\end{equation}

\begin{equation}
g'(\lambda) = \begin{cases}
    1 & \text{if $\lambda > \theta$} \\
    \exp\left(\lambda/\theta - 1 \right) & \text{otherwise}
    \end{cases}
\end{equation}

\begin{equation}
g''(\lambda) = \begin{cases}
    0 & \text{if $\lambda > \theta$} \\
    \dfrac {1} {\theta} \exp\left(\lambda/\theta - 1 \right) & \text{otherwise}
    \end{cases}
\end{equation}

g is continuous, has continous gradients and is convex since the second derivative is positive almost everywhere and the gradient is continous where the second derivative is undefined.

The derivative of g is always between 0 and 1. For this reason
\begin{equation}
    0 \le (g(x)-g(y)) / (x-y) \le 1    
    \label{eq:K_is_contraction_for_our_choice}
\end{equation}

For this reason when backpropagating through this function the gradient magnitude w.r.t. Frobenius norm is guaranteed to decrease, since we do a componentwise multiplication with values between 0 and 1. Therefore if the gradient with respect to A is bounded then the gradient with respect to B will be bounded too. since the mapping from network output to B preserves norms this means that the gradient with respect to the network output is bounded as well.

\subsection{Study of $-log(|A|)$}
Here we show that the term $-log(|A|)$ has the properties we desire.

\begin{equation}
    \dfrac
        {\partial -log(|A|)}
        {\partial A} =
    {A^{-1}}^T = A^{-1}
    \label{eq:grad_norm_term}
\end{equation}

The first step follows from Bishop Appendix C\cite{bishop2006pattern}.
The second step comes from the fact that A is symmetric.

\textbf{Bounded gradients}
Let D and V be the eigenvalue decomposition of B.

By using equation \ref{eq:diag_remapping_theorem} we get
\begin{align}
||\dfrac{\partial log(|A|)} {\partial B}||_F &= ||V^T(f(D)^{-1} \circ K(D, r))V||_F \\ &=
\sum\limits_{i=0}^d 1/r(\lambda_i) \circ r'(\lambda_i) \\&\le \sum\limits_{i=0}^d 1/\theta \\ &= \dfrac {d} {\theta}
\end{align}

\textbf{Convexity}
We will show that the method is convex when all eigenvalues are larger than $\theta$. i.e. when $g$ is an identity mapping.

Since this part of the loss does not depend on $\nu$ it is sufficient to prove that the loss is convex w.r.t. A.

For this part we will use a flattening function $f: \mathbb{R}^{d^2} \rightarrow \mathbb{R}^{d\times d}$ such that $f(v)_{n,m} = v_{d*(n-1)+m}$.

We will study $-log(|f(a)|)$ and prove its convexity w.r.t. v. we will use $f(a) = A$ to simplify notation.

\begin{align}
     H_{d*(i-1)+j,d*(k-1)+l} &= \dfrac
         {\partial^2 -log(|f(a)|)}
         {\partial f(a)_{i, j} \partial f(a)_{k, l}} \\
     &= \dfrac
         {\partial {f(a)^{-1}}_{i, j}}
         {\partial f(a)_{k, l}} \\
     &= (A^{-1} \dfrac {\partial A} {\partial A_{k,l}} A^{-1})_{i,j} \\
     &= \mathbf{e}_i^T A^{-1} \textbf{e}_k \textbf{e}_l^T A^{-1}\textbf{e}_j \\
     &= A^{-1}_{i, k}A^{-1}_{l, j}
\end{align}

We will show that this matrix is postive definite.
Shorthand $f(x) = X$ note X needs to be positive definite.

\begin{align}
    \bx^TH\bx &= 
    \sum\limits_{n,m = 0}^d \sum\limits_{i,j =0}^d \bx_{d(n-1)+m} H_{d(n-1)+m, d(i-1)+j} \bx_{d(i-1)+j} \\ &=
    \sum\limits_{n,m = 0}^d \sum\limits_{i,j =0}^d X_{n,m} A^{-1}_{i,n} A^{-1}_{j,m} X_{i,j} \\ &=
    \sum\limits_{m = 0}^d{XA^{-1}XA^{-1}}_{m,m} \\ &=
    tr(XA^{-1}XA^{-1})  \\ &=
    tr(U^{-1}\hat{D}UU^{-1}\hat{D}U) \\ &=
    tr(\hat{D}\hat{D}U^{-1}U)  \\ &=
    \sum\limits_{n=0}^d \hat{D}_{i,i}^2 \ge 0
\end{align}

We define $U$ and $\hat{D}$ by $U^T\hat{D}U = XA^{-1}$ such that $U$ is ON and $\hat{D}$ is diagonal.

The last step relies on the fact that the eigenvalues of $XA^{-1}$ are real. We will show this in the following lemma.

\paragraph{Lemma: Eigenvalues for multiplication of real symmetric matrices.} This lemma and proof is very similar to the discussion here \cite{spectrum_of_symm_real_mult}. 
For two symmetric real matrices $A$ and $B$ where $A$ is also positive definite then the eigenvalues of $AB$ are real.

\paragraph{Proof}
Since A is symmetric and real there exist an eigenvalue decomposition $A = V^TDV$. Where D is diagonal, real with an inverse while V is ON.
Then $AB = V^TDV B$ Then reparameterize B as $B = V^TXV$. X will still be symmetric ($X=VBV^T=VB^TV^T=VV^TX^TVV^T = X^T$)
Therefore $AB = V^TDXV$, since a basis change does not change the eigenvalues $AB$ will have the same eigenvalues as $DX$.

Assume $d$ and $\mathbf{v}$ is a pair of eigenvalues and eigenvectors of $DX$.

\begin{equation}
    d \mathbf{v}^{*} D^{-1} \mathbf{v} = \\
    \mathbf{v}^{*} D^{-1}DX \mathbf{v} = \\
    \mathbf{v}^{*} X \mathbf{v} = \\
    \mathbf{v}^{*} X^{*} \mathbf{v} = \\
    \mathbf{v}^{*} X^{*} D^{*} D^{-1} \mathbf{v} = \\
    d^{*} \mathbf{v}^{*} D^{-1} \mathbf{v}
\end{equation}

Step 1 is based on $d \mathbf{v} = DX \mathbf{v}$.
Step 2 is based on $D^{-1}D = I$. Note that $D^{-1}$ exists since A is positive definite.
Step 3 is based on $X^{*} = X^{T} = X$ since X is real and symmetric.
Step 4 is based on $D^{-1}D = I$ and $D^{*}=D$ since D is real.
Step 5 is done by $\mathbf{v}^{*}X^{*}D^{*} = (DX\mathbf{v})^{*}$ since $D$ and therefore $D^{-1}$ is positive definite we know that $\mathbf{v}^{*} D^{-1} \mathbf{v} = \sum\limits_{i=0}^d |\mathbf{v}_i|^2 / D_{i,i} > 0$. Since all $D_{i,i} > 0$. If we divide the first and last expression by this number we get $d = d^{*}$ and therefore d is real. This concludes the proof $\qedsymbol$.

We use the previous lemma and conclude that our function is convex when the remapping is an identity mapping, i.e. for the set where all eigenvalues of A are larger than $\theta$.

\paragraph{Bounded Hessians:} If $||\mathbf{x}||_2 = 1$ then $||X||_F = 1$ and then $||XA^{-1}||_F \le ||X||_F||A^{-1}||_F = ||A^{-1}||_F$
\begin{equation}
    tr(XA^{-1}XA^{-1}) = <(XA^{-1})^T, XA^{-1}>_F \le \\ ||XA^{-1}||_F^2 \le ||A^{-1}||_F^2 \le \dfrac {d} {\theta^2}
\end{equation}

Where $<. , .>_F$ is the Frobenius inner product.

We have now showed that this part of the loss has bounded gradients everywhere and that it is convex with bounded Hessians where eigenvalues are larger than $\theta$.

\subsection{Study of $h_1(||A\mathbf{x}-\nu||_2)$}

In this section we show that $h(||A\mathbf{x}-\nu||_2)$ has the desired properties. i.e. convex respect to $A$ and $\nu$ in the region where all eigenvalues of $A$ are larger than $\theta$, bounded Hessians for the same region and bounded gradients.

\subsubsection{Properties in region $||A\mathbf{x}-\nu||_2 < 1$}
Here we will show that we have the desired properties in this region.
If $||A\bx-\nu||_2 < 1$ then this term is

\begin{equation}
    J = \dfrac {(A\bx-\nu)^T(A\bx-\nu)} {2} = \dfrac {\bx^TA^TA\bx -2\nu^TA\bx + \nu^T\nu} {2}
\end{equation}

We have

\begin{equation}
    \dfrac {\partial J} {\partial \nu_i} = \nu_i - (A\bx)_i \implies
    \dfrac {\partial J} {\partial \nu} = (\nu - A\bx)
\end{equation}

\begin{equation}
    \dfrac {\partial^2 J} {\partial \nu_i \partial \nu_j} = I(i=j)
\end{equation}

\begin{equation}
    \dfrac {\partial^2 J} {\partial \nu_i \partial A_{k,l}} = I(k=i) \bx_l
\end{equation}

\begin{equation}
    \dfrac {\partial J} {\partial A_{i,j}} = \bx_j(A\bx-\nu)_i
\end{equation}

\begin{equation}
    \dfrac {\partial^2 J} {\partial A_{i,j} \partial A_{k,l}} = I(k=i) \bx_j \bx_l
\end{equation}

We now know the Hessian, we will use a flattening function $f: \mathbb{R}^d \times \mathbb{R}^{d\times d} \rightarrow \mathbb{R}^{d+d^2}$
\begin{equation}
    f(\mathbf{c}, B)_{i} = \begin{cases}
        \mathbf{c}_i \text{ if } i \le d \\
        B_{\lfloor i/d\rfloor, ((i)\%d)+1} \text{ otherwise}
    \end{cases}
\end{equation}
Where $\%$ indicates the remainder function.

\begin{align}
    &f(\mathbf{c}, B)^T H f(\mathbf{c}, B) \\ &= 
    \sum\limits_{(i,j)} I(i=j) \mathbf{c}_i^T\mathbf{c}_i +
    2 \sum\limits_{(i,k,l)} \mathbf{c}_i I(k=i) \bx_l B_{k,l} +
    \sum\limits_{i,j,k,l} B_{i,j}B_{k,l} I(k=i) \bx_j \bx_l \\ &=
    \textbf{c}^T\mathbf{c} + 2\textbf{c}^TB\bx + \bx^TB^TB\bx \\
     &= ||(B\bx+\textbf{c})||_2^2 \ge 0
\end{align}

Therefore the function is convex in this region. By maximizing $B$ and $\mathbf{c}$ such that $||f(B,\mathbf{c})||_2 = 1$ we find that the 2 norm of H is $(||\bx||_2^2+1)$. We can compute the Frobenius norm from its definition and sum and realize that $||H||_F = \sqrt{d*(||\bx||_2^4 + ||\bx||_2^2 + 1)} = \sqrt{d}(||\bx||_2^2+1)$ Where $d$ is the dimensionality of $\bx$.

In this region the gradients are bounded by
\begin{equation}
    \sqrt{||\nu-A\bx||_2^2 + ||\bx||_2^2||A\bx-\nu||_2^2} < \\
    \sqrt{||\bx||_2^2+1}
\end{equation}

\subsubsection{Properties in region $||A\bx-\nu||_2 > 1$}

For this region the term turns into

$J = \sqrt{(A\bx-\nu)^T(A\bx-\nu)} -1/2$

We will now compute gradients and Hessians.

\begin{equation}
    \dfrac {\partial J} {\partial \nu_i} =
    \dfrac
        {(A\bx-\nu)_i}
        {\sqrt{(A\bx-\nu)^T(A\bx-\nu)}}
\end{equation}

\begin{align}
    \dfrac {\partial^2 J} {\partial \nu_i \partial \nu_j} &=
    \dfrac
        {I(i=j)\sqrt{(A\bx-\nu)^T(A\bx-\nu)} - 
             \dfrac
                 {(A\bx-\nu)_i(A\bx-\nu)_j)}
                 {\sqrt{(A\bx-\nu)^T(A\bx-\nu)}}
        }
        {((A\bx-\nu)^T(A\bx-\nu))^{3/2}} \\ &=
    \dfrac
        {I(i=j)(A\bx-\nu)^T(A\bx-\nu) - 
            (A\bx-\nu)_i(A\bx-\nu)_j
        }
        {((A\bx-\nu)^T(A\bx-\nu))^{2}}
\end{align}

\begin{align}
    \dfrac {\partial^2 J} {\partial \nu_i \partial A_{k,l}} &= 
    \dfrac
        {I(k=i)\bx_l\sqrt{(A\bx-\nu)^T(A\bx-\nu)} -
        \dfrac
            {(A\bx-\nu)_i \bx_l(A\bx-\nu)_k}
            {\sqrt{(A\bx-\nu)^T(A\bx-\nu)}}}
        {((A\bx-\nu)^T(A\bx-\nu))^{3/2}} \\ &=
    \bx_l \dfrac
        {I(k=i)(A\bx-\nu)^T(A\bx-\nu) -
             (A\bx-\nu)_i (A\bx-\nu)_k}
        {((A\bx-\nu)^T(A\bx-\nu))^2}
\end{align}

\begin{equation}
    \dfrac {\partial J} {\partial A_{i,j}} =
    \dfrac
        {\bx_j(A\bx-\nu)_i}
        {\sqrt{(A\bx-\nu)^T(A\bx-\nu)}}
\end{equation}

\begin{align}
    \dfrac {\partial^2 J} {\partial A_{i,j} \partial A_{k,l}} &=
    \bx_j\bx_l\dfrac
        {I(i=k)\sqrt{(A\bx-\nu)^T(A\bx-\nu)} -
        \dfrac
           {(A\bx-\nu)_i(A\bx-\nu)_k}
           {\sqrt{(A\bx-\nu)^T(A\bx-\nu)}}
        }
        {((A\bx-\nu)^T(A\bx-\nu))^{3/2}} \\ &=
    \bx_j\bx_l\dfrac
        {I(i=k)(A\bx-\nu)^T(A\bx-\nu) -
            (A\bx-\nu)_i(A\bx-\nu)_k
        }
        {((A\bx-\nu)^T(A\bx-\nu))^{2}}
\end{align}

Now we see that the norm of the gradient is $\sqrt{||\bx||_2^2+1}$

We use the flattening function again
\begin{align}
    &f(\textbf{c}, B)^THf(\textbf{c},B) = \\ &= 
    \dfrac 
        1
       {((A\bx-\nu)^T(A\bx-\nu))^{2}}
    (\sum\limits_{i,j} \textbf{c}_i\textbf{c}_j (I(i=j)||A\bx-\nu||_2^2 \\ &-
         (A\bx-\nu)_i(A\bx-\nu)_j) \\ &+
         2\sum\limits_{i,k,l} \textbf{c}_i B_{k,l}(I(i=k)\bx_l(A\bx-\nu)^T(A\bx-\nu) -(A\bx-\nu)_i \bx_l(A\bx-\nu)_k)  \\ &+
         \sum\limits_{i,j,k,l} B_{i,j}B_{k,l} \bx_l\bx_j(I(k=i)||A\bx-\nu||_2^2 - (A\bx-\nu)_i(A\bx-\nu)_k))
     \\ &=
    \dfrac
        {||\textbf{c}||_2^2+2\textbf{c}^TB\bx+x^TB^TB\bx}
        {((A\bx-\nu)^T(A\bx-\nu))}
    \\ &-
    \dfrac
          {
          (\textbf{c}^T(A\bx-\nu))^2 + 2\textbf{c}^T(A\bx-\nu)(A\bx-\nu)^TB\bx + \bx^TB^T(A\bx-\nu)(A\bx-\nu)B\bx}
          {((A\bx-\nu)^T(A\bx-\nu))^2} \\ &=
    \dfrac
        {||\textbf{c}+B\bx||_2^2}
        {||(A\bx-\nu)||_2^2}
    - \dfrac
        {((\textbf{c}+B\bx)^T(A\bx-\nu))^2}
        {||(A\bx-\nu)||_2^4} \\ &\ge
    \dfrac
        {||\textbf{c}+B\bx||_2^2}
        {||(A\bx-\nu)||_2^2}
    - \dfrac
        {||\textbf{c}+B\bx||_2^2||A\bx-\nu||_2^2}
        {||(A\bx-\nu)||_2^4} \\
        &= 0
\end{align}

The second to last step is from Cauchys inequality. This concludes the proof that the function has a positive semidefinite Hessian in both regions.

We can also notice that the Hessian has eigenvalues of magnitude less than $(1+||\bx||_2^2) / ||A\bx-\nu||_2^2 \le (1+||\bx||_2^2)$ The Frobenius norm of the Hessian is $\sqrt{d-1}(1+||\bx||_2^2)$

Finally we notice that $h(\bx)$ is continous with continous gradients. Therefore $h(||A\bx-\nu||)$ will also be continous with continous gradients w.r.t $\nu$ and $A$.

If we consider two points $(A_1, \nu_1)$ and $(A_2, \nu_2)$ and consider the function
$p(\lambda) = h(||(\lambda A_1 + (1-\lambda) A_2) \bx - (\lambda \nu_1 + (1- \lambda) \nu_2)||_2)$
Then this function will have a positive second derivative almost everwhere and at the place where the second derviative is undefined the deriative is continous. Therefore this function is convex. Therefore our function is convex for every line segment. Therefore the function is convex for the convex set where all eigenvalues of A are larger than $\theta$.

\newpage
\section{Extra tables}
The numbers we report for this section are based on running the same experiment 5 times with different random seeds. The  number we report is the mean of these runs. The value after the $\pm$ sign is the empirical standard deviation of these 5 runs.
\label{sec:supp_extra_tables}
\begin{table}[htbp]
\caption{Ablation of different pretraining datasets for WFLW. The bottom two rows indicate that using pretrained imagenet weights give significant improvements over random initialization. The top two lines indicate that pretraining on a face dataset gives a small improvement in performance.
All runs in this table use a resnet101 backbone with a convolution instead of average pooling at the end. The loss used is our Huber loss with $\nu$ parameterization.}
\centering
\bobcaptionspace
\begin{tabular}{lccr}
\toprule
Pretrain dataset & epochs & NME & NLL \\
\midrule
 300W-LP\cite{zhu2016face} & 200 & 4.70 $\pm$ 0.03 & -344.6 $\pm$ 1.60 \\
ImageNet & 200 & 4.76 $\pm$ 0.06 & -355.4 $\pm$  3.30 \\
ImageNet & 50 & 4.91 $\pm$ 0.04 & -355.8 $\pm$ 0.79 \\
None & 50 & 5.31 $\pm$ 0.04 & -342.2 $\pm$ 1.20 \\
\bottomrule
\end{tabular}
\end{table}
\begin{table}[htbp]
\caption{Ablation for test time augmentation (TTA) for WFLW. Using probabilistic TTA significantly improves performance compared to no TTA. All runs use resnet101 backbone with convolution instead of average pooling at the end. The methods were trained for 200 epochs. We use our loss with the $\nu$ parameterization.}
\centering
\bobcaptionspace
\begin{tabular}{lcr}
\toprule
Pretraining dataset & TTA & NME \\
\midrule
300W-LP\cite{zhu2016face} & \checkmark & 4.58 $\pm$ 0.02 \\
300W-LP\cite{zhu2016face} & \xmark & 4.70 $\pm$ 0.03 \\
ImageNet & \checkmark & 4.62 $\pm$ 0.04 \\
ImageNet & \xmark & 4.76 $\pm$ 0.06 \\
\bottomrule
\end{tabular}
\end{table}
\begin{table}[htbp]
\caption{Comparison when training for 200 epochs compared to 50 on WFLW. When training for longer the estimated position performance continues to increase for longer than the NLL. We use resnet101 backbone and our loss with $\nu$ parameterization. Models are pretrained on Imagenet.}
\centering
\bobcaptionspace
\begin{tabular}{lcr}
\toprule
Epochs & NME & NLL \\
\midrule
200 & 4.76 $\pm$   0.06 & -355.4 $\pm$  3.30 \\
50 & 4.91 $\pm$ 0.04 & -355.8 $\pm$ 0.79 \\
\bottomrule
\end{tabular}
\end{table}

\begin{table}[htbp]
\caption{Ablation of fusion type for mpii. Performance difference is small and probably not significant. Models trained for 50 epochs using resnet101 as backbone with our loss using a $\nu$ parameterization.}
\centering
\bobcaptionspace
\begin{tabular}{lr}
\toprule
Fusion type & PCKh@0.5 \\
\midrule
probabilistic & 85.0 $\pm$ 0.1 \\
mean & 84.8 $\pm$ 0.1 \\ 
\bottomrule
\end{tabular}
\end{table}

\begin{table}[htbp]
  \centering
      \caption{Network architecture ablation for WFLW. ResNet18 performs worse than the other two architectures. ResNet101 and ResNet50 has similar performance.}
     \bobcaptionspace
    \begin{tabular}{lcc}
    Network & NME ($\downarrow$) & NLL ($\downarrow$) \\
      \midrule
         ResNet101 & 4.91 $\pm$ 0.04 & -355.8 $\pm$ 0.79 \\
         ResNet50 & 4.89 $\pm$ 0.02 & -357.2 $\pm$ 0.47\\
         ResNet18 & 5.01 $\pm$ 0.02 & -351.0 $\pm$ 0.75\\
         \bottomrule
    \end{tabular}
    \label{tab:WFLW arch_ablation}
\end{table}

\begin{table}[h!]
\caption{Comparison average pooling at end versus using channelwise convolutions. Experiment shows that using channelwise convolutions instead of average pooling significantly improve performance. Models use resnet101 backbone, trained for 50 epochs using our loss with $\nu$ parameterization.}
\centering
\bobcaptionspace
\begin{tabular}{ccr}
\toprule
Average pooling at end & NME & NLL \\
\midrule
\xmark & 4.91 $\pm$ 0.04 & -355.8 $\pm$ 0.79 \\
\checkmark & 5.25 $\pm$ 0.02 & -336.5 $\pm$ 0.40  \\
\bottomrule
\end{tabular}
\end{table}
\newpage
\section{MLE of multiple multivariate Huber distribution predictions}
\label{sec:supp_huber_fusion}

For many applications there will be multiple estimates of the target position. For example one could have multiple views of a person and with our approach it would be possible to generate a multivariate Huber distribution from each view, creating multiple estimates of each landmark. Unfortunately, the Huber distribution is not closed under multiplication, unlike the normal distributions. However, we have created an efficient method which is based on the majorize/minimize method for quadratic functions. For the special case $\delta = 0$ this method would turn into Weiszfeld's algorithm \cite{kuhn1973note}.

We want to find the maximum likelihood point given $n$ independent multi-variate Huber distributions. 
Let each independent estimate of $\by$ be parameterized by $(\nu_i, A_i)$ then
\begin{equation}
p(\by) \propto \prod\limits_{i=1}^n \exp\left(-h_{\delta}(\|A_i \by - \nu_i\|)\right) 
\end{equation}
and the optimal $\by$ is found from:
\begin{equation}
\underset{\by}{\arg\max}\; p(\by) = \underset{\by}{\arg \min} \sum\limits_{i=1}^n h_{\delta}(\|A_i \by - \nu_i\|) = \underset{\by}{\arg\min} \sum_{i=1}^n g_i(\by)
\end{equation}
This optimization problem can be solved with a Majorize-Minimization (MM) procedure. If $\by^{(t)}$ is the current estimate for the optimal $\by$ then a tight quadratic majorizer for each $g_i(\by)$ is:
\begin{align}
  q_i(\by \mid \by^{(t)}) =
  \begin{cases}
    \|A_i \by - \nu_i\|^2 / 2 & \text{if $\|A_i \by^{(t)} - \nu_i\| < \delta$}\\[4pt]
    \frac{\delta \|A_i \by - \nu_i\|^2}{2\|A_i \by^{(t)} - \nu_i\|} + \dfrac {\delta\|A_i \by^{(t)} - \nu_i\| -\delta^2} {2} & \text{otherwise}
  \end{cases}
\end{align}
It is then simple to majorize $\sum_{i=1}^n g_i(\by)$ with
\begin{align}
  q(\by \mid \by^{(t)}) = \sum_{i=1}^n q_i(\by \mid \by^{(t)})
\end{align}
By iteratively solving $\by^{(t+1)} = \arg\min_{\by} q(\by \mid \by^{(t)})$, we converge to the desired maximum likelihood estimate solution. Since $q(\by|\by^{(t)})$ is a quadratic function with respect to $\by$ finding the minima for each step is easy.

\setcounter{section}{6}
%\pgfplotsset{colormap={mycolormap}{
%        rgb255=(0,0,0)
%        rgb255=(255,255,255)
%    },
%}

\section{Visualizations of $L_2$ multivariate Huber pdf}
\label{sec:huber_vis}
This section presents visualizations of the $L_2$ multivariate Huber distribution to aid understanding the effect of the parameters on the shape, spread and effective support of the distribution. The $\Lambda$ parameter plays a similar role in the shape of the distribution as in a Gaussian distribution. The $\delta$ parameter controls the tail behaviour of the distribution and its spread given the orientation defined by $\Lambda$. Crucially, the parameters $\delta$ and $\Lambda$ can be independently set to change the spread of the distribution. This means that even when $\delta$ is kept fixed one can still adapt the distribution's support via $\Lambda$ to down-weight outliers in our loss. A less drastic change in $\Lambda$ is needed for our Huber distribution, given a reasonable value of $\delta$, to adapt to outliers than for a Gaussian distribution.

In the following figures it is assumed each distributions shown has zero mean vector. Each plot shows the iso-probability contours of the distribution marking the $.005, .05, .2, .35, .5, .65,$ $.8, .95$ and $.99$ percentiles. The shading in each ring is proportional to log of the mean probability of the distribution in that region. The scaling - applied to the spatial and shading components - is constant across the plots within a figure.

%%%%%%%%%%%%%%%%%%%%%%%%%%%%%%%%%%%%%%%%%%%%%%%%
%% E[X^2] constant = Sigma, delta changes
\def\cwid{0.2pt}
\def\pwid{.15\linewidth}
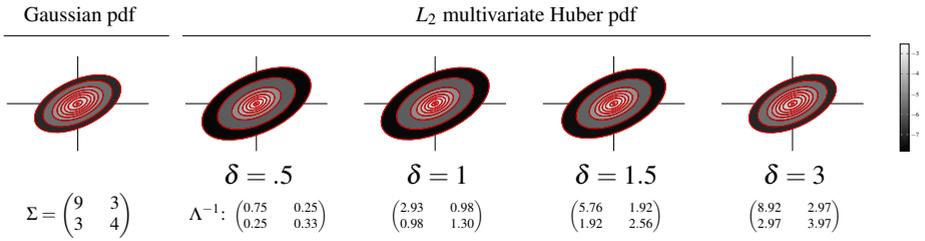
\begin{figure}[ht]
\begin{center}
\begin{tabular}{cccccc}
   {\footnotesize Gaussian pdf}&
   \multicolumn{4}{c}{{\footnotesize $L_2$ multivariate Huber pdf}}\\
   \cmidrule(ll){1-1}
  \cmidrule(rl){2-5}
  %% Gauss
    \resizebox{\pwid}{!}{%
      \begin{tikzpicture}
        \draw (-1.5,0) -- (1.5,0) (0,-1) -- (0,1);
        \foreach \x / \y / \col in {0.979 / 0.4889 / 0.1565, 0.78965 / 0.39434 / 0.42589, 0.57869 / 0.28899 / 0.57466, 0.4674 / 0.23341 / 0.65684, 0.37966 / 0.1896 / 0.71387, 0.29934 / 0.14949 / 0.75757, 0.21548 / 0.10761 / 0.79304, 0.10322 / 0.051548 / 0.81298, 0.032257 / 0.016109 / 0.81776}{
          \definecolor{mycolor}{rgb}{\col, \col, \col};
          \draw[rotate =-154.9007,double=red,ultra thin,double distance=\cwid,fill=mycolor] (0,0) ellipse ({\x} and {\y});
        }
      \end{tikzpicture}
    }%
    &
    \resizebox{\pwid}{!}{%
      \begin{tikzpicture}
        \draw (-1.5,0) -- (1.5,0) (0,-1) -- (0,1);
        %% delta = .5
        \foreach \x / \y / \col in {1.2361 / 0.61727 / 0.026193, 0.88334 / 0.44113 / 0.3601, 0.55762 / 0.27846 / 0.54764, 0.41323 / 0.20636 / 0.66469, 0.31269 / 0.15615 / 0.75528, 0.23012 / 0.11492 / 0.83519, 0.15369 / 0.076752 / 0.91924, 0.06656 / 0.033239 / 0.97908, 0.019922 / 0.0099485 / 1}{
          \definecolor{mycolor}{rgb}{\col, \col, \col};
          \draw[rotate =-154.9007,double=red,ultra thin,double distance=\cwid,fill=mycolor] (0,0) ellipse ({\x} and {\y});
        }        
      \end{tikzpicture}
    }%
    &
    \resizebox{\pwid}{!}{%
\begin{tikzpicture}
  \draw (-1.5,0) -- (1.5,0) (0,-1) -- (0,1);  
  %% delta = 1
  \foreach \x / \y / \col in {1.2263 / 0.61239 / 0.029899, 0.87815 / 0.43854 / 0.3632, 0.55705 / 0.27818 / 0.55032, 0.41489 / 0.20719 / 0.66676, 0.31632 / 0.15796 / 0.75643, 0.23577 / 0.11774 / 0.83416, 0.16184 / 0.080819 / 0.89856, 0.075218 / 0.037563 / 0.9323, 0.023356 / 0.011664 / 0.94005}{
    \definecolor{mycolor}{rgb}{\col, \col, \col};
    \draw[rotate =-154.9007,double=red,ultra thin,double distance=\cwid,fill=mycolor] (0,0) ellipse ({\x} and {\y});
  }  
\end{tikzpicture}
    }%
    &
    \resizebox{\pwid}{!}{%
\begin{tikzpicture}
  \draw (-1.5,0) -- (1.5,0) (0,-1) -- (0,1);  
  %% delta = 1.5
  \foreach \x / \y / \col in {1.1776 / 0.58809 / 0.049195, 0.85348 / 0.42621 / 0.37963, 0.55591 / 0.27761 / 0.56409, 0.42584 / 0.21266 / 0.67538, 0.33628 / 0.16793 / 0.75123, 0.26118 / 0.13043 / 0.80563, 0.18608 / 0.092924 / 0.84784, 0.088522 / 0.044207 / 0.87094, 0.027615 / 0.01379 / 0.87642}{
    \definecolor{mycolor}{rgb}{\col, \col, \col};
    \draw[rotate =-154.9007,double=red,ultra thin,double distance=\cwid,fill=mycolor] (0,0) ellipse ({\x} and {\y});
  }  
\end{tikzpicture}
    }%
    &
    \resizebox{\pwid}{!}{%
\begin{tikzpicture}
  \draw (-1.5,0) -- (1.5,0) (0,-1) -- (0,1);  
  %% delta = 3
  \foreach \x / \y / \col in {0.98839 / 0.49359 / 0.15057, 0.78924 / 0.39413 / 0.4256, 0.57723 / 0.28826 / 0.5753, 0.46609 / 0.23276 / 0.65783, 0.3784 / 0.18897 / 0.71503, 0.29841 / 0.14902 / 0.75877, 0.2149 / 0.10732 / 0.79434, 0.10279 / 0.051332 / 0.81433, 0.032122 / 0.016041 / 0.81911}{
    \definecolor{mycolor}{rgb}{\col, \col, \col};
    \draw[rotate =-154.9007,double=red,ultra thin,double distance=\cwid,fill=mycolor] (0,0) ellipse ({\x} and {\y});
  }  
\end{tikzpicture}
    }%
&\resizebox{.015\textheight}{!}{%
\begin{tikzpicture}
  \pgfplotscolorbardrawstandalone[colormap name=mycolormap, point meta min=-7.822,point meta max=-2.522,colorbar style={xtick={-8,-6,-4,-2}}]
\end{tikzpicture}
    }%
    \\
    &
    $\delta = .5$ &
    $\delta = 1$ &
    $\delta = 1.5$ &
    $\delta = 3$\\    
    {\scriptsize $\Sigma = \begin{pmatrix} 9 & 3 \\ 3 & 4\end{pmatrix}$} &

    {\scriptsize $\Lambda^{-1}$:}    
    %\delta=.5
    {\tiny $\begin{pmatrix*}[r] 0.75 &  0.25 \\  0.25  & 0.33 \end{pmatrix*}$} &
    %\delta = 1.0
    {\tiny $\begin{pmatrix*}[r] 2.93  & 0.98 \\ 0.98 & 1.30 \end{pmatrix*}$} &
    % \delta = 1.5
    {\tiny $\begin{pmatrix*}[r] 5.76  &  1.92 \\ 1.92 &  2.56 \end{pmatrix*}$} &
    %\delta = 3
    {\tiny $\begin{pmatrix*}[r] 8.92 &  2.97 \\ 2.97 &   3.97 \end{pmatrix*}$}        
   %%  {\scriptsize $\Lambda$:}    
   %%  {\tiny $\begin{pmatrix*}[r] 1.78 & -1.33\\ -1.33 & 4.00 \end{pmatrix*}$} &
   %% {\tiny $\begin{pmatrix*}[r] 0.46 & -0.34 \\ -0.34  & 1.03	   \end{pmatrix*}$} &
   %% {\tiny $\begin{pmatrix*}[r] 0.23 & -0.17 \\ -0.17 & 0.52 \end{pmatrix*}$} &
   %% {\tiny $\begin{pmatrix*}[r] 0.15 &  -0.11\\ -0.11 &   0.34 \end{pmatrix*}$}    
\end{tabular}
\end{center}
\bobcaptionspace
\caption[]{\textbf{Comparison of $L_2$ Huber distributions and a bivariate Gaussian distribution all with the same second-order moment matrix}. 
  \textbf{Leftmost plot}: The bivariate normal distribution whose covariance matrix, $\Sigma$, by definition equals $\text{E}[\mathbf{X}^2]$. \textbf{Other plots}: Each plot shows a $L_2$ multivariate Huber distribution whose second order moment matrix equals that of the distribution shown in the leftmost plot. For a Huber distribution $\text{E}[\mathbf{X}^2] = \alpha(\delta) \Lambda^{-1}$. The parameters defining the shown Huber distributions are given under the plot. As $\delta$ increases: 1) The parameter matrix $\Lambda^{-1}$ changes, entries increase in magnitude,  to keep $\text{E}[\mathbf{X}^2]$ fixed and the spread of distribution decreases. 2) The distribution increasingly resembles a Gaussian distribution. 3) Less of the probability mass of the pdf is contained in the tails.}
\label{fig:huber_same_cov}
\end{figure}
%%%%%%%%%%%%%%%%%%%%%%%%%%%%%%%%%%%%%%%%%%%%%%%%

%%%%%%%%%%%%%%%%%%%%%%%%%%%%%%%%%%%%%%%%%%%%%%%%%%%%%%%%%%%%%%%
%% Sigma is fixed, delta changes, log-scale
\def\pwid{.145\linewidth}
\begin{figure}[htpb]
\begin{center}
\begin{tabular}{cccccc}
    {\small Guassian pdf} & \multicolumn{4}{c}{$L_2$ multivariate Huber pdf with $\Lambda =
      ${\scriptsize $\begin{pmatrix} 9 & 3 \\ 3 &
          4\end{pmatrix}^{-1}$}}\\ \cmidrule(ll){1-1} \cmidrule(ll){2-5}
  %% Gauss
    \resizebox{\pwid}{!}{%
      \begin{tikzpicture}
        \draw (-2.5,0) -- (2.5,0) (0,-1.7) -- (0,1.7);
        %\draw (-1.5,0) -- (1.5,0) (0,-1) -- (0,1);
        \foreach \x / \y / \col in {0.979 / 0.4889 / 0.41589, 0.78965 / 0.39434 / 0.65385, 0.57869 / 0.28899 / 0.78527, 0.4674 / 0.23341 / 0.85786, 0.37966 / 0.1896 / 0.90824, 0.29934 / 0.14949 / 0.94684, 0.21548 / 0.10761 / 0.97817, 0.10322 / 0.051548 / 0.99578, 0.032257 / 0.016109 / 1}{
          \definecolor{mycolor}{rgb}{\col, \col, \col};
          \draw[rotate =-154.9007,double=red,ultra thin,double distance=\cwid,fill=mycolor] (0,0) ellipse ({\x} and {\y});
        }
      \end{tikzpicture}
    }%
    &
    \resizebox{\pwid}{!}{%
      \begin{tikzpicture}
        \draw (-2.5,0) -- (2.5,0) (0,-1.7) -- (0,1.7);
        %% delta = .8
        \foreach \x / \y / \col in {2.6802 / 1.3385 / 0.042949, 1.9167 / 0.95717 / 0.33778, 1.2116 / 0.60504 / 0.5033, 0.89932 / 0.44911 / 0.60656, 0.68191 / 0.34054 / 0.68633, 0.50385 / 0.25162 / 0.75655, 0.33934 / 0.16946 / 0.82463, 0.15322 / 0.076516 / 0.86397, 0.047095 / 0.023519 / 0.87315}{
          \definecolor{mycolor}{rgb}{\col, \col, \col};
          \draw[rotate =-154.9007,double=red,ultra thin,double distance=\cwid,fill=mycolor] (0,0) ellipse ({\x} and {\y});
        }        
      \end{tikzpicture}
    }%
    &
    \resizebox{\pwid}{!}{%
\begin{tikzpicture}
  \draw (-2.5,0) -- (2.5,0) (0,-1.7) -- (0,1.7);
  %% delta = 1
  \foreach \x / \y / \col in {2.1509 / 1.0741 / 0.11676, 1.5403 / 0.76919 / 0.41118, 0.97706 / 0.48793 / 0.57647, 0.72772 / 0.36341 / 0.67932, 0.55482 / 0.27707 / 0.75854, 0.41353 / 0.20651 / 0.8272, 0.28386 / 0.14176 / 0.88408, 0.13193 / 0.065884 / 0.91388, 0.040966 / 0.020458 / 0.92073}{
    \definecolor{mycolor}{rgb}{\col, \col, \col};
    \draw[rotate =-154.9007,double=red,ultra thin,double distance=\cwid,fill=mycolor] (0,0) ellipse ({\x} and {\y});
  }  
\end{tikzpicture}
    }%
    &
    \resizebox{\pwid}{!}{%
      \begin{tikzpicture}
        \draw (-2.5,0) -- (2.5,0) (0,-1.7) -- (0,1.7);
  %% delta = 1.5
  \foreach \x / \y / \col in {1.4719 / 0.73504 / 0.24676, 1.0667 / 0.53271 / 0.53864, 0.69482 / 0.34698 / 0.70159, 0.53224 / 0.26579 / 0.79989, 0.42031 / 0.2099 / 0.86689, 0.32644 / 0.16302 / 0.91494, 0.23257 / 0.11614 / 0.95223, 0.11064 / 0.055253 / 0.97263, 0.034515 / 0.017236 / 0.97748}{
    \definecolor{mycolor}{rgb}{\col, \col, \col};
    \draw[rotate =-154.9007,double=red,ultra thin,double distance=\cwid,fill=mycolor] (0,0) ellipse ({\x} and {\y});
  }  
\end{tikzpicture}
    }%
    &
    \resizebox{\pwid}{!}{%
\begin{tikzpicture}
  \draw (-2.5,0) -- (2.5,0) (0,-1.7) -- (0,1.7);  
  %% delta = 3
  \foreach \x / \y / \col in {0.99255 / 0.49566 / 0.40926, 0.79255 / 0.39579 / 0.6522, 0.57966 / 0.28947 / 0.78443, 0.46805 / 0.23374 / 0.85734, 0.37999 / 0.18976 / 0.90786, 0.29967 / 0.14965 / 0.9465, 0.2158 / 0.10777 / 0.97792, 0.10322 / 0.051548 / 0.99557, 0.032257 / 0.016109 / 0.9998}{
    \definecolor{mycolor}{rgb}{\col, \col, \col};
    \draw[rotate =-154.9007,double=red,ultra thin,double distance=\cwid,fill=mycolor] (0,0) ellipse ({\x} and {\y});
  }  
\end{tikzpicture}
    }%
&\resizebox{.015\textheight}{!}{%
\begin{tikzpicture}
  \pgfplotscolorbardrawstandalone[colormap name=mycolormap, point meta min=-9.49,point meta max=-3.49,colorbar style={xtick={-8,-6,-4,-2}}]
\end{tikzpicture}
    }%
    \\
    (a) {\scriptsize $\Sigma = \begin{pmatrix} 9 & 3
        \\ 3 & 4\end{pmatrix}$}    & (b) $\delta = .8$ & (c) $\delta = 1$ & (d) $\delta = 1.5$ &
    (e) $\delta = 3$
    %% \\ {\scriptsize $\Sigma = \begin{pmatrix} 9 & 3
    %%     \\ 3 & 4\end{pmatrix}$}    
\end{tabular}
\end{center}
\bobcaptionspace
  \caption[]{\textbf{Comparison of $L_2$ Huber distributions with the
      same $\Lambda$ but different $\delta$ parameter}.
    (a) The bivariate normal distribution with covariance matrix $\Sigma$. (b-e) Each plot shows a $L_2$ multivariate Huber distribution with the same $\Lambda$ parameter but different $\delta$.  As $\delta$ increases: 1) The spread of distribution decreases. 2) The distribution increasingly resembles a Gaussian distribution. 3) Less of the probability mass of the pdf is contained in the tails.}
\label{fig:huber_delta_varies}  
\end{figure}
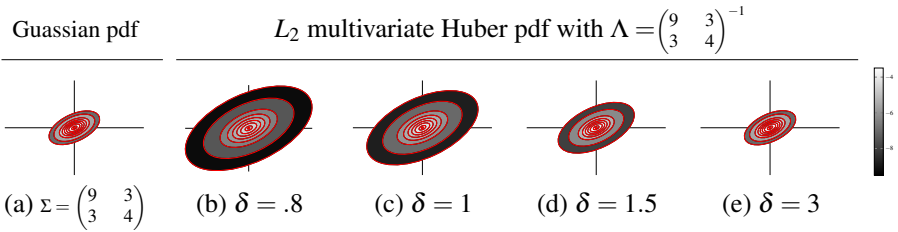

%%%%%%%%%%%%%%%%%%%%%%%%%%%%%%%%%%%%%%%%%%%%%%%%%%%%%%%%%
%% Sigma varies, delta fixed, log-scale
\def\pwid{.25\linewidth}
\begin{figure}
\begin{center}
\begin{tabular}{cccc}
  %%delta = 1, alpha = .5
    \resizebox{\pwid}{!}{%
      \begin{tikzpicture}
        \draw (-3,0) -- (3,0) (0,-2) -- (0,2.0);
        %\draw (-1.5,0) -- (1.5,0) (0,-1) -- (0,1);
        \foreach \x / \y / \col in {0.75952 / 1.5209 / 0.23229, 0.5439 / 1.0891 / 0.52671, 0.34502 / 0.69089 / 0.692, 0.25697 / 0.51457 / 0.79484, 0.19592 / 0.39232 / 0.87406, 0.14603 / 0.29241 / 0.94272, 0.10024 / 0.20072 / 0.9996, 0.046587 / 0.093289 / 1.0294, 0.014466 / 0.028968 / 1.0363}{
          \definecolor{mycolor}{rgb}{\col, \col, \col};
          \draw[rotate =115.0972,double=red,ultra thin,double distance=\cwid,fill=mycolor] (0,0) ellipse ({\x} and {\y});
        }
      \end{tikzpicture}
    }%
    &
      %%delta = 1, alpha = 1
      \resizebox{\pwid}{!}{%
      \begin{tikzpicture}
        \draw (-3,0) -- (3,0) (0,-2.0) -- (0,2.0);
        %\draw (-1.5,0) -- (1.5,0) (0,-1) -- (0,1);
        \foreach \x / \y / \col in {1.0741 / 2.1509 / 0.11676, 0.76919 / 1.5403 / 0.41118, 0.48793 / 0.97706 / 0.57647, 0.36341 / 0.72772 / 0.67932, 0.27707 / 0.55482 / 0.75854, 0.20651 / 0.41353 / 0.8272, 0.14176 / 0.28386 / 0.88408, 0.065884 / 0.13193 / 0.91388, 0.020458 / 0.040966 / 0.92073}{
          \definecolor{mycolor}{rgb}{\col, \col, \col};
          \draw[rotate =115.0972,double=red,ultra thin,double distance=\cwid,fill=mycolor] (0,0) ellipse ({\x} and {\y});
        }
      \end{tikzpicture}
      }%
    &
      %%delta = 1, alpha = 2
      \resizebox{\pwid}{!}{%
      \begin{tikzpicture}
        \draw (-3.0,0) -- (3.0,0) (0,-2.0) -- (0,2.0);
        %\draw (-1.5,0) -- (1.5,0) (0,-1) -- (0,1);
        \foreach \x / \y / \col in {1.519 / 3.0418 / 0.0012389, 1.0878 / 2.1783 / 0.29566, 0.69004 / 1.3818 / 0.46095, 0.51394 / 1.0291 / 0.5638, 0.39183 / 0.78463 / 0.64301, 0.29205 / 0.58483 / 0.71167, 0.20047 / 0.40144 / 0.76856, 0.093174 / 0.18658 / 0.79836, 0.028932 / 0.057935 / 0.80521}{
          \definecolor{mycolor}{rgb}{\col, \col, \col};
          \draw[rotate =115.0972,double=red,ultra thin,double distance=\cwid,fill=mycolor] (0,0) ellipse ({\x} and {\y});
        }
      \end{tikzpicture}
      }%
      &%\resizebox{.015\textheight}{!}{%
\resizebox{.02\textheight}{!}{%        
\begin{tikzpicture}
  \pgfplotscolorbardrawstandalone[colormap name=mycolormap, point meta min=-9.49,point meta max=-3.49,colorbar style={xtick={-8,-6,-4,-2}}]
\end{tikzpicture}
    }%      
    \\
    (a) {\small $\Lambda = \Lambda_1  / .5$} & (b) {\small $\Lambda = \Lambda_1$}
    & (c) {\small $\Lambda = \Lambda_1 / 2$}
\end{tabular}
\end{center}
\bobcaptionspace
  \caption[]{\textbf{Comparison of $L_2$ Huber distributions with the
      same $\delta=1$ parameter but different $\Lambda$}.
    By decreasing the magnitude of values in $\Lambda_1 = ${\tiny $\begin{pmatrix*} 9 &3\\ 3 & 4\end{pmatrix*}^{-1}$} by the same factor one can increase the spread of the distribution.}
\label{fig:huber_Lambda_varies}    
\end{figure}
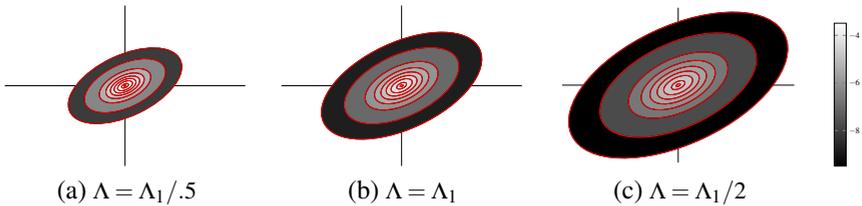

%%%%%%%%%%%%%%%%%%%%%%%%%%%%%%%%%%%%%%%%%%%%%%%%
.

\end{document}